\documentclass[lettersize,journal]{IEEEtran}
\usepackage{algorithm,algorithmic}
\usepackage{array}
\usepackage[caption=false,font=footnotesize,labelfont=rm,textfont=rm]{subfig}
\usepackage{textcomp}
\usepackage{stfloats}
\usepackage{url}
\usepackage{verbatim}

\usepackage{graphicx}
\usepackage{cite}
\usepackage{booktabs}
\usepackage{multirow}
\usepackage{epstopdf}
\usepackage{hyperref}
\usepackage{graphicx}
\usepackage{amsmath, amsthm, amsfonts}
\usepackage{bbm}

\usepackage{soul}
\usepackage{color, xcolor}
\usepackage{siunitx} 
\begin{document}


\title{Bridging Predictive Uncertainty and Safe Action: Sample-Conditioned Differentiable Planning for Autonomous Driving}

\author{
   Chengzhen Meng, Pei Liu, Zhiyu Huang, Chen Lv, \textit{Senior Member, IEEE}, and Jun Ma, \textit{Senior Member, IEEE}
    \thanks{Chengzhen Meng, Pei Liu, and Jun Ma are with the Robotics and Autonomous Systems Thrust, The Hong Kong University of Science and Technology, Guangzhou, China, and also the Department of Electronic and Computer Engineering, The Hong Kong University of Science and Technology, Hong Kong SAR, China (e-mail: cmeng403@connect.hkust-gz.edu.cn; pliu061@connect.hkust-gz.edu.cn; jun.ma@ust.hk). }
    \thanks{Zhiyu Huang is with the Department of Civil and Environmental Engineering, University of California, Los Angeles, CA 90095 USA (e-mail: zhiyuh@ucla.edu).} 
    \thanks{Chen Lv is with the School of Mechanical and Aerospace Engineering, Nanyang Technological University, Singapore 639798 (e-mail: lyuchen@ntu.edu.sg).} 
	}




\maketitle
\begin{abstract}

Complex, dynamic, and interactive driving environments pose significant challenges for autonomous driving, primarily due to the pervasive uncertainty of surrounding traffic. A fundamental bottleneck in current systems is the disconnect between highly expressive uncertainty modeling and interpretable, safe motion planning. In this paper, we propose a novel sample-conditioned differentiable planning framework that bridges this gap by explicitly incorporating diffusion-generated future trajectories into the optimization process. Rather than compressing predictions into a single deterministic future or relying on black-box end-to-end architectures, our approach leverages a conditional diffusion model to generate a diverse set of plausible future scenarios. Crucially, these samples are directly fed into a differentiable planner, which explicitly mitigates predictive uncertainty via an empirical Conditional Value-at-Risk (CVaR) tail-risk constraint. This allows the planner to optimize a physically interpretable trajectory that is robust to rare yet safety-critical interactions. Furthermore, we introduce a directed graph representation for scene context that yields substantial improvements in both predictive effectiveness and computational efficiency. Validated through extensive open-loop and closed-loop evaluations on the Waymo Open Motion and Argoverse 2 datasets, our framework significantly outperforms state-of-the-art baselines in safety, efficiency, and ride comfort. 

\end{abstract}

\begin{IEEEkeywords} 
Autonomous driving, diffusion model, differentiable planning, uncertainty-aware planning
\end{IEEEkeywords}

\section{Introduction}\label{sec:intro}
\IEEEPARstart{N}{avigating} complex, interactive traffic environments remains a fundamental challenge for autonomous driving. The primary difficulty stems from the pervasive uncertainty inherent in the future behaviors of surrounding agents (SAs) \cite{10122777,11077782,10122127}. To ensure safety without resorting to overly conservative driving, an autonomous vehicle must not only accurately anticipate these uncertain futures but also explicitly reason over them during motion planning.

\begin{figure}[t]
    \includegraphics[width=1.0\linewidth]{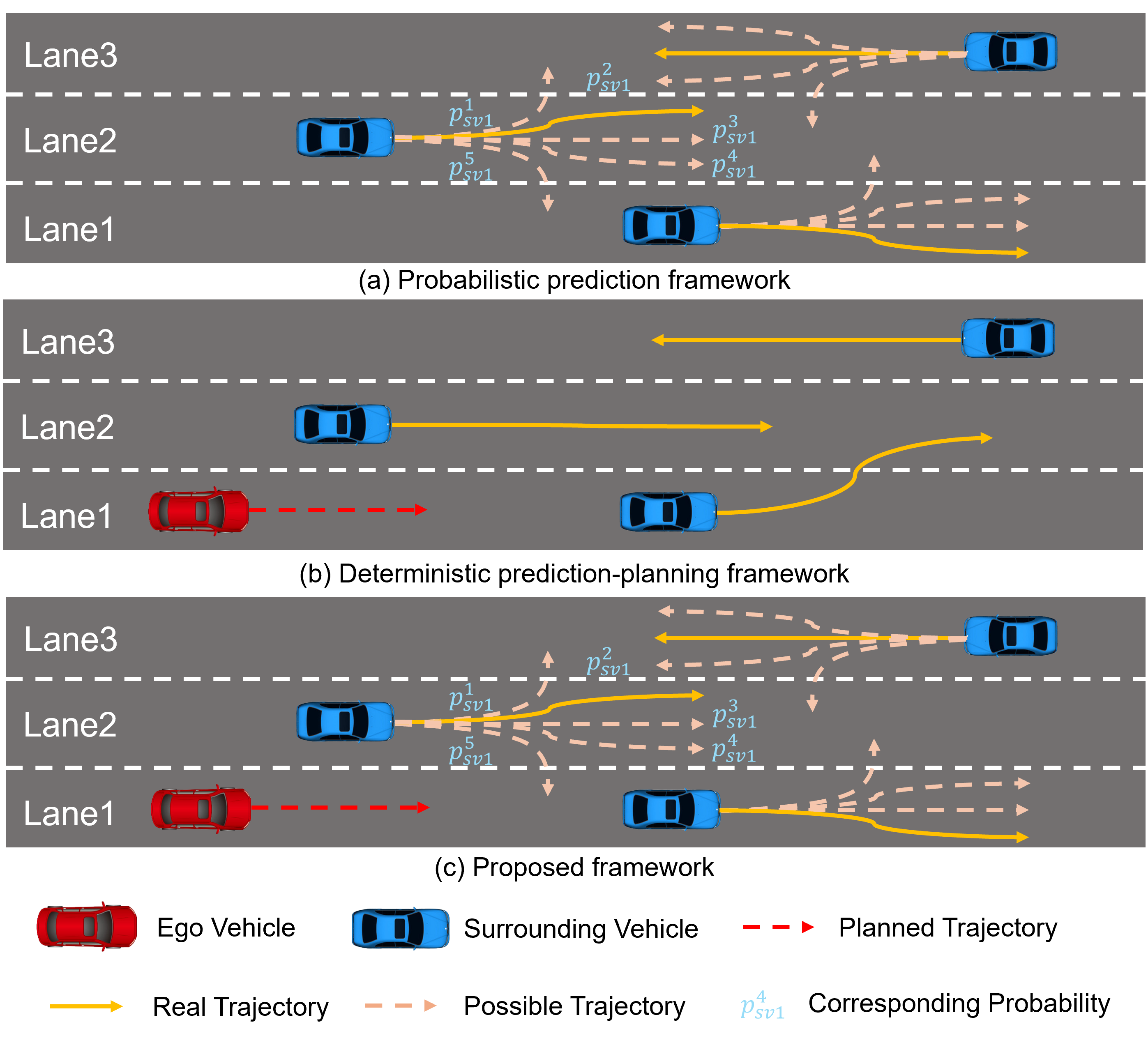}
   \caption{Learning-based frameworks for Autonomous Driving. (a) Probabilistic prediction frameworks without downstream planning. (b) Deterministic prediction-planning frameworks that optimize against single-point estimates. (c) Our proposed framework that integrates probabilistic prediction with differentiable uncertainty-aware optimization.
   }
    \label{fig1}
\end{figure}

Traditional approaches attempt to handle this uncertainty by explicitly modeling environmental dynamics and deriving safe boundaries through mathematical optimization or rule-based logic \cite{noh2017decision, sankar2020adaptive,11081470,11494278}. For instance, probabilistic reachable sets \cite{10952911} or chance-constrained model predictive control (MPC) \cite{9145612} have been employed to guarantee safety margins under uncertainty. However, these model-based methods heavily rely on restrictive analytical assumptions, such as Gaussian noise perturbations \cite{10952911} or unimodal/non-Gaussian mixture distributions \cite{9145612}, which rarely hold in highly dynamic real-world traffic \cite{Schwarting_2018, Salzmann2020}. Consequently, their capacity to represent complex uncertainty is severely limited.

Conversely, learning-based methods excel at capturing complex distributions directly from massive driving data \cite{Lee_2017_CVPR,Gupta_2018_CVPR,Chai2018CoRL,pmlr-v155-zhao21b}. While earlier models often struggle to stably fit complex continuous distributions \cite{9756903,Rhinehart_2018,8953435}, recent diffusion-based generative models (Fig.~\ref{fig1}(a)) have demonstrated remarkable capability in modeling the joint distribution of multi-agent trajectories, generating highly realistic and diverse future scenarios \cite{Gu_2022_CVPR, Song_2021_ICLR, Jiang_2023_CVPR}. Despite this success in the prediction stage, a critical disconnect remains: how to effectively propagate this rich uncertainty into the planning stage. While some recent works attempt to bypass this by employing end-to-end architectures that directly decode ego-vehicle plans from sensor inputs or latent representations \cite{Jiang_2023_CVPR,Bo_2026_ICLR}, they inherently suffer from a black-box nature. The implicit encoding of planning decisions within deep neural networks lacks physical interpretability, making it exceedingly difficult to enforce strict safety constraints or post-hoc verification—an unacceptable risk for safety-critical applications \cite{goodfellow2014explaining,rudin2019stop}.

To overcome the opacity of pure black-box models, an alternative line of research seeks to restore interpretability by embedding explicit differentiable optimization modules into the learning pipeline \cite{9197560,10740461,liu2024integrating} (Fig.~\ref{fig1}(b)). For example, differentiable integrated prediction and planning (DIPP) frameworks learn planning cost functions directly from data, enabling joint training \cite{huang2023differentiable}. However, a critical limitation persists: these frameworks typically compress predicted futures into a single deterministic estimate before planning. Consequently, the rich uncertainty information contained in multiple plausible future behaviors is discarded. This deterministic bottleneck severely compromises safety in highly interactive scenarios, where avoiding collisions depends on anticipating rare yet risky evolutions.

To break these bottlenecks, we propose a sample-conditioned differentiable planning framework (Fig.~\ref{fig1}(c)) that seamlessly integrates the expressive power of diffusion models with the rigorous safety guarantees of optimization-based planning. Instead of collapsing predictions, our framework explicitly conditions trajectory optimization on a diverse set of diffusion-generated future samples. To handle the combinatorial challenge of optimizing against multiple discrete samples, we formulate the planning process as a differentiable optimization problem constrained by an empirical Conditional Value-at-Risk (CVaR). This tail-risk formulation forces the planner to explicitly account for highly hazardous future interactions. Benefiting from a fully differentiable design, the diffusion predictor and the optimization-based planner are trained jointly end-to-end. This framework allows the planner to directly exploit the structured uncertainty of the predictive distribution while maintaining a transparent, physically interpretable decision logic.
The main contributions are summarized as follows:

\begin{itemize}
 \item \textbf{A Novel Sample-Conditioned Planning Framework:} We propose a fully differentiable framework that explicitly incorporates diffusion-generated future trajectories of SAs into downstream planning, enabling the ego vehicle (EV) to reason over uncertain traffic behaviors of SAs rather than relying on deterministic predictions.

\item \textbf{Risk-Aware Differentiable Optimization:} We formulate motion planning as a sample-based optimization problem and introduce an empirical CVaR tail-risk safety constraint. This provides a mathematically tractable way to penalize severe safety violations across sampled futures within a differentiable pipeline.

\item \textbf{Efficient Directed Graph Representation}: We introduce a directed graph structure to encode global scene context, replacing conventional vectorized maps. This representation explicitly models topological relationships, significantly reducing computational overhead while enhancing the model's ability to capture complex agent-scene interactions.

\item \textbf{Comprehensive Empirical Validation:} Extensive open-loop and closed-loop experiments on the Waymo Open Motion \cite{ettinger2021large} and Argoverse 2 datasets \cite{wilsonargoverse} demonstrate that our jointly trained framework achieves state-of-the-art performance. Specifically, it significantly outperforms existing baselines in safety, planning accuracy, and driving comfort, demonstrating the superiority of explicitly coupling diffusion-based uncertainty with differentiable planning.
\end{itemize}

The remainder of this paper is organized as follows. Sec.~\ref{sec:formu} details the problem formulation. Sec.~\ref{sec:method} elaborates on the proposed framework, comprising the conditional diffusion-based predictor and the sample-conditioned differentiable planner. Sec.~\ref{sec:exp} presents comprehensive open-loop and closed-loop evaluations, along with ablation studies and discussions on limitations. Finally, Sec.~\ref{sec:cond} concludes the paper.

\section{Problem Formulation}\label{sec:formu}
Unlike conventional deterministic planning methods that optimize trajectories based on a single predicted future, sample-conditioned planning refers to a planning paradigm in which trajectory optimization is explicitly conditioned on a set of sampled future trajectories representing the predictive distribution of SAs. Given historical observations and map contexts, the prediction module models SAs' futures as a conditional trajectory distribution. Multiple future trajectory samples are then generated and incorporated into an uncertainty-aware planning optimization process, enabling the planner to reason over uncertainty during trajectory generation. 

We define the agents set as $\mathcal{N}=\left \{0,1,...,N \right \}$, where $0$ denotes the EV and $1,...,N$ denotes the SAs. We also define the state set $S=\left \{ S_{0},S_1,...,S_N \right \}$ and action set $A=\left\{A_{0},A_1,...,A_N \right \}$. The state of each agent is defined as $S_i=\left[x,y,\theta,v_x,v_y\right]$, where $x,y$ represent the location of the agent, $\theta$ represents the heading angle of the agent, and $v_x,v_y$ represent the velocity of the agent in the $x$ and $y$ direction, respectively. The actions of each agent are defined as the state variation $A_i=\left[\Delta x,\Delta y,\Delta \theta,\Delta v_x,\Delta v_y\right]$. We define the current timestep as $t=0$, given the historical state of all agents for the previous $H$ timesteps $S_{0,1,...,N}^{-H:0}$ and the scene context, the predictor needs to predict $M$ possible joint future actions of SAs for the next $T$ timesteps $\hat{A}=\left \{ \hat{A}_{i,m}|m=1,2,...,M\right\}, \hat{A}_i=\hat{A}_{1,...,N}^{1:T}$ and output an initial control sequence $\hat{u}^{1:T}$ of EV, where $\hat{A}_{i}^{1:T}$ is the predicted action of agent $i$ for the next timesteps $T$. We also define the probability of each future $P=\left \{ p_m|m=1,2,...,M\right\}$. With the current state $S_{1,...,N}^{0}$, we can accumulate the states of each agent over the future time period $T$, $\hat{S}=\left \{ \hat{S}_{i,m}|m=1,2,...,M\right\},\hat{S}_i=\hat{S}_{1,...,N}^{1:T}$, where $\hat{S}_{i}^{t}$ is the predicted states of agent $i$ at timestep $t$. 

Rather than selecting a single predicted future, the planner explicitly conditions trajectory optimization on the entire set of sampled future trajectories, the initial control sequence of the EV $\hat{u}^{1:T}$, and the cost function. We formulate the motion planning problem as a nonlinear least squares problem, where the planning horizon is the same as the prediction horizon $T$ and the optimization variables are control variables of EV $\textbf{u}^{1:T}=\left \{\textbf{u}^1,\textbf{u}^2,...,\textbf{u}^{T}\right \}$. The objective function includes multiple factors considered during driving, including driving comfort, driving efficiency, and safety. We denote each cost as $j_i$ and the corresponding weight as $\omega_i$.

\section{Sample-Conditioned Differentiable Planning with Diffusion-Based Prediction}
\label{sec:method}
\begin{figure*}[ht]
    \includegraphics[width=0.95\linewidth]{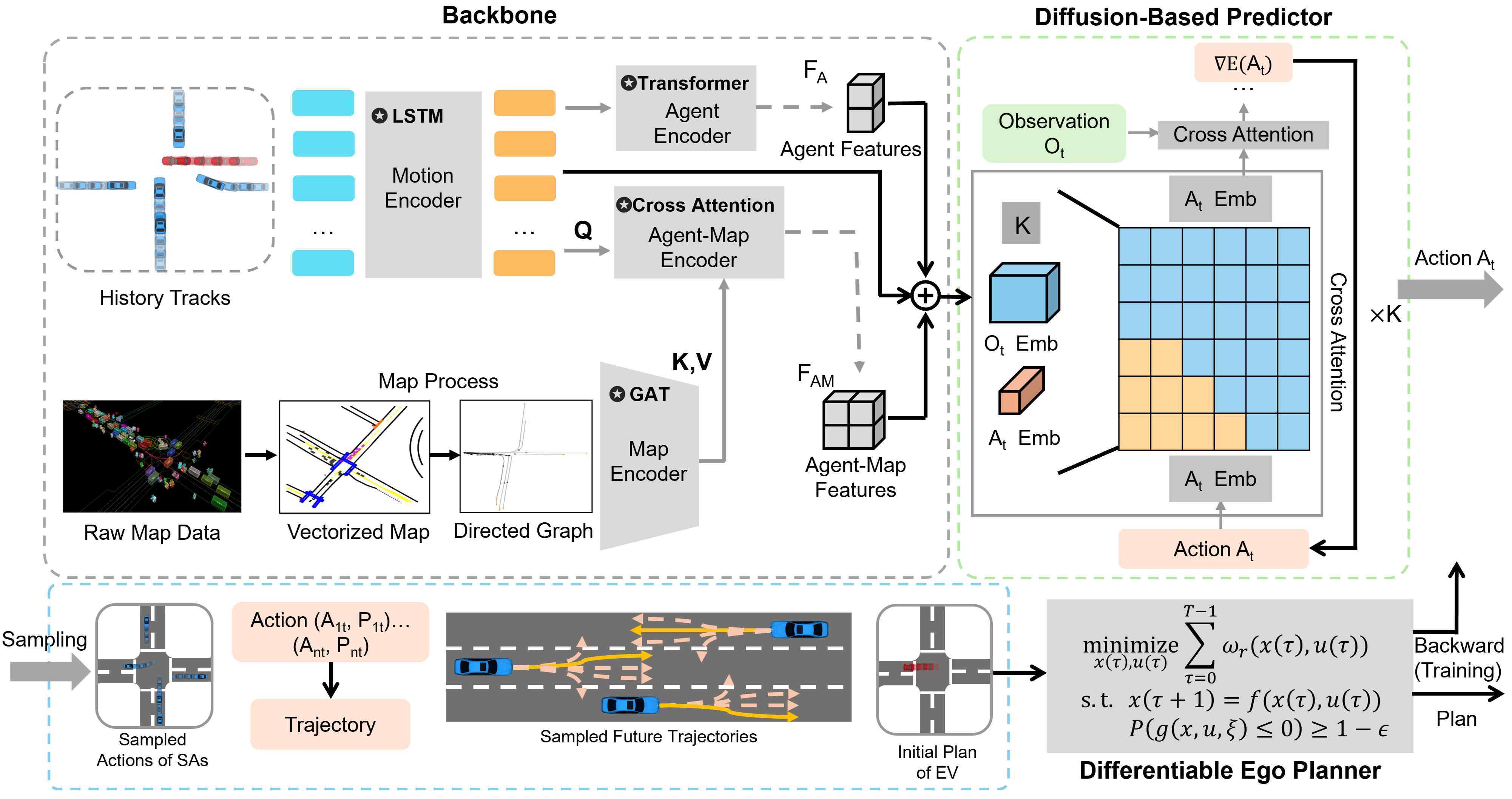}
   \caption{The overall architecture of our proposed framework. The model first encodes agent histories and global scene context into a concatenated representation. A conditional diffusion model then leverages these embeddings to sample future trajectories for surrounding agents. Finally, these samples are fed into a differentiable planner, which optimizes a safe ego trajectory by explicitly mitigating predictive uncertainty via a CVaR-based chance constraint.}
    \label{fig2}
\end{figure*}

In this section, we introduce the proposed framework, which mainly contains three parts, as shown in Fig.~\ref{fig2}. First, the input historical states together with scene context undergo processing and are projected into a high-dimensional representation space. These resulting embeddings subsequently serve as queries, keys, and values (Q, K, and V) within an attention module, from which agent-level features as well as agent-scene interaction features are derived. The concatenation of these two types of features and the motion features is used as the observation embeddings in a transformer-based diffusion network, which directly generates multiple possible trajectories of all agents under the conditional probability distribution. Second, we employ a differentiable optimizer as a motion planner to explicitly plan a future trajectory for the EV according to all possible predicted trajectories and the initial plan, which is obtained by the predicted trajectory of the EV. Specifically, we formulate the motion planning problem as a nonlinear least squares problem and use the Gauss-Newton method to solve the problem. Since the optimizer is differentiable, gradients computed by the planner can be backpropagated to the prediction module and cost function during training, which makes the overall framework end-to-end trainable. The detailed introduction of the diffusion-based predictor and the sample-conditioned motion planner is given in Section \ref{diff} and Section \ref{planner}.
\subsection{Diffusion-Based Prediction}
\label{diff}
\subsubsection{Input Representation}
The proposed framework takes two categories of inputs: historical agent states and scene context. For every agent, its historical trajectory comprises dynamic features spanning the previous $H = 20$ time steps, denoted as $S_{0,1,...,N}^{-20:0}$, which encompasses 2-D coordinates, orientation, and speed. We focus on the $N = 10$ agents closest to EV, with their historical observations organized in a fixed-dimension tensor where absent agents are zero-padded. Regarding scene context, two map component types are incorporated: lane segments and pedestrian crossings. All positional attributes for both agents and map components are transformed into the EV-centric local coordinate frame.
\subsubsection{Concatenated Observation Embedding}
To create the concatenated observation embedding required by the diffusion-based predictor, the input historical agent states and scene context are processed in three steps. The first step is \textbf{encoding agents' history and scene context}. Specifically, each agent's state trajectory is processed by a long short-term memory (LSTM) encoder to distill its historical motion pattern. Compared to Transformers, the LSTM offers preferable end-task prediction quality and greater efficiency for short-horizon sequential data, which justifies its selection. The encoded outputs from this recurrent network are taken as the motion features. Finally, these historical state encodings for the full set of agents (with the EV among them) are organized into a tensor format.

For scene context encoding, the scene context encoder comprises a lane encoder and a crosswalk encoder to process lane and crosswalk vectors. Following the data format of \cite{huang2023differentiable}, we first transform raw inputs into a vectorized map representing each agent's local scene context, including traversable lanes and surrounding crosswalks. Lane waypoint features encompass positions and headings of lane centers and boundaries, while crosswalk waypoint features consist of points delineating crosswalk regions. All positional information is transformed into the EV's local coordinate system. However, this vectorized map format has its drawbacks. First, there is a lack of connection between the local scene context of each agent. The local context for each agent only shows the lane it might take and the surrounding crosswalks, making it difficult to represent the global scene features, which may result in the omission of key scene features. Second, the number of agents can be quite large, especially when extending to large-scale scenes, leading to an excessive amount of local scene context. In \cite{huang2023differentiable}, the local scene context of each agent is encoded and then stacked, resulting in an overwhelming number of features that significantly increases the training time required. 

\begin{figure}[t]
\includegraphics[width=0.95\linewidth]{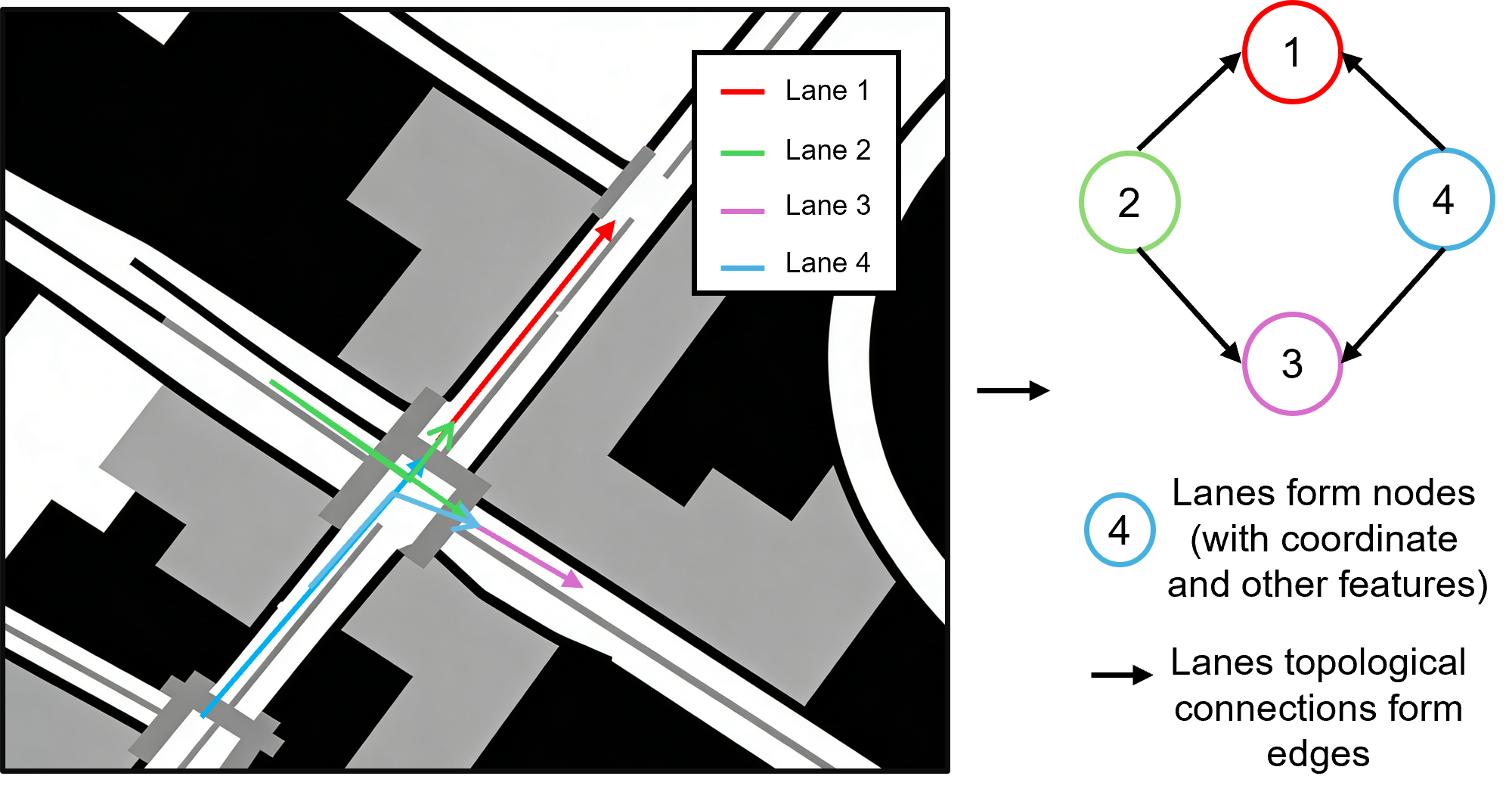}
   \caption{Construction of the directed graph representation from a standard vectorized map. In this formulation, lane segments serve as graph nodes, while edges are established based on driving directions and traffic rules to capture global topological context.}
    \label{fig3}
\end{figure}

So in our work, for the lane encoder, we propose a directed graph transformation method. Lane segments from local contexts serve as directed graph nodes, with edges established based on driving directions and traffic regulations to model lane connectivity, as shown in Fig.~\ref{fig3}. This approach ensures global context fusion while explicitly modeling topological relationships. Since all agents' contexts can be represented through this unified global scene context, the dimensionality of subsequent processing is reduced, accelerating training. The processed lane graph is encoded via graph attention networks (GAT) as part of the global scene context. For crosswalk elements, we collect the complete set of crosswalk data and utilize a multilayer perceptron (MLP) to project the numerical descriptors (position and heading) into an embedding space. The resulting features are incorporated as an extra part of the holistic scene context, to be leveraged for agent-scene interaction encoding in the following stages.

The second step of creating the concatenated observation embedding is \textbf{modeling agent and agent–map features}. Modeling the interdependencies and correlations between agents and their surrounding environment plays a critical role in maintaining accurate predictions. We follow the idea of \cite{huang2023differentiable} and extract features of agents and agents to map. The agent–agent interaction encoder is implemented as a two-layer self-attention Transformer, using the last timestep of the encoded motion features as Q, K, and V to model how agents influence one another. 

With each agent's encoded motion features, two multi-head cross-attention encoders are applied to model agent-scene interactions: one layer encodes the interactions between agents and lanes, and another layer encodes the interactions between agents and crosswalks. We employ the last timestep of the encoded motion features as queries (Q), with encoded lane information or encoded crosswalk information serving as keys and values (K and V). This operation is invoked twice to process agent-lane and agent-crosswalk interactions separately, yielding attention features for agent-lane vectors and agent-crosswalk vectors. These two vectors are then concatenated along the feature dimension to form the agent-scene context attention features. Subsequently, we introduce a multi-head cross-attention mechanism where the last timestep of the agents' encoded motion features remains as queries, and scene context features serve as keys and values, thereby generating encodings of agent-scene context relationships. We employ this encoder for each agent individually to capture how they interact with the environmental context, subsequently stacking the output features along the axis corresponding to future timesteps.

Finally, we concatenate motion features, agent interaction features, and agent-scene interaction features along the feature dimension to form the concatenated observation embedding $O_t$, which is subsequently fed into the diffusion model. Since motion features retain the temporal dimension, while the other two interaction features adopt the last timestep of motion features and thus lack temporal information, we expand the temporal dimension for both interaction features before concatenation with motion features. The resulting observation embedding $O_t$ preserves temporal dynamics while capturing both agent-to-agent and agent-to-map interactions.

\subsubsection{Denoising Diffusion Probabilistic Models for Prediction}
DDPMs are generative models that synthesize samples via iterative denoising based on Langevin dynamics. 
In our task, we use a DDPM to approximate the conditional distribution $p(A_t|O_t)$, which allows the model to predict actions conditioned on observations without the cost of inferring future states, speeding up the diffusion process and improving the accuracy of generated actions \cite{diff1}. To sample $A_{k-1}$, we follow the equation:
\begin{equation}
    A_{k-1,t}=\alpha (A_{k,t}-\gamma\varepsilon _\theta (O_t,A_{k,t},k)+\mathcal{N}(0,\sigma ^2I))
,
\label{equa:diff2}
\end{equation}
where $t$ is the timestep and $k$ is the iteration number. Here, $\varepsilon _{\theta (x_k,k)}$ represents the noise prediction network with parameters $\theta$ and $\mathcal{N}(0,\sigma ^2I)$ represents Gaussian noise added at each iteration. $O_t$ contains high-dimensional features of history tracks and scene context. Equation (\ref{equa:diff2}) can be converted to a single noisy gradient descent step:
\begin{equation}
    A_{k,t}\gets A_{k,t}-\gamma\bigtriangledown E(O_t,A_{k,t},k)
,
\label{equa:diff3}
\end{equation}
where the noise prediction network $\varepsilon _\theta (O_t,A_{k,t},k)$ effectively predicts the gradient field $\bigtriangledown E(O_t,A_{k,t},k)$, which approximates the action-score gradient, i.e., $\bigtriangledown\log p_\theta (A_t|O_t)$ \cite{song2019generative}, and $\gamma$ is the learning rate. By learning the parameters of the action-score gradient and performing Stochastic Langevin Dynamics sampling on this gradient field to simulate the distribution $p(A_t|O_t)$, which means it can simulate various uncertain actions that a vehicle might exhibit under the current observation during operation.

We adopt the transformer-based diffusion architecture from \cite{diff1}, which uses the minGPT \cite{brown2020language} transformer decoder model for action prediction, and modify the inputs as follows. 
The transformer-based diffusion architecture receives the concatenated observation embedding $O_t$, which compactly represents all agents' historical states and the scene context. Noise-infused actions are tokenized and streamed into the transformer decoder. Both the $k$-th diffusion step inputs and the index $k$ are sinusoidally encoded, with $k$ serving as the prepended token. Before entering the decoder, observations $O_t$ are flattened into a sequence and augmented with positional embeddings. During the $k$-th iteration, the decoder estimates the noise to be removed at each timestep of the action sequence. A causal attention mask ensures that each action token attends only to itself and prior tokens. The model ultimately outputs $T$ diffusion-denoised action steps and an initial EV control sequence $\hat{u}^{1:T}$, the latter obtained by feeding the predicted EV action sequence $\hat{A}_0$ into an MLP. To stabilize the diffusion process \cite{salimans2022progressive}, the model directly outputs $\hat{A}_{0,1,...,N}$ instead of predicting noise.

\subsection{Sample-Conditioned Differentiable Planning}
\label{planner}
\subsubsection{Planning Problem Formulation}
We formulate the planning problem as an open-loop, finite-horizon optimal control problem, aiming to determine a control sequence $u^*=\{u_1,u_2,...,u_T\}$ that minimizes a composite cost function. The cost function is constructed as a weighted sum of squared terms, where each term corresponds to a distinct loss component $j_i$ multiplied by its associated weighting coefficient $\omega_i$, as follows:

\begin{equation}
    J= \frac{1}{2} \sum_i{\left \| \omega _ij_i(u,\hat{S}  ) \right \|^2 },
    \label{cost}
\end{equation}
where $j_i$ denotes distinct cost functions, each of which depends on the control inputs $u$ and the predicted states of SAs $\hat{S}$. The predicted states of SAs encompass $M$ possible realizations, generated by the preceding prediction module.

With the control variable $u_t=(a_t,\delta_t)$, we adopt the kinematic bicycle model \cite{rajamani2011lateral} shown in (\ref{kmodel}) to update the states of the EV:

\begin{equation}
\begin{cases}
x_{t+1}=x_t+v_t\text{cos}(\theta_t)\Delta  t, \\
y_{t+1}=y_t+v_t\text{sin}(\theta_t)\Delta  t, \\
\theta_{t+1}=\theta_t+\frac{v_t}{L}\text{tan}(\delta_t)\Delta  t,\\
v_{t+1}=v_t+a_t\Delta  t, 
\label{kmodel}
\end{cases}
\end{equation}
where $L$ is the wheelbase of the vehicle and $\Delta  t$ is the time interval. The overall planning problem can be formulated as follows:

\begin{equation}
\begin{aligned}
\label{model}
 \min_{u} & \quad \frac{1}{2}\sum_i {\left \| \omega _ij_i(u,\hat{S}  ) \right \|^2 }, \\
 \text{s.t.} &  \quad S_0^{t+1}=f(u_t,S_0^{t}).
\end{aligned}
\end{equation}
During training, the weights $\omega _i$ associated with these cost functions serve as learnable parameters. Gradients of the ultimate planning error can be backpropagated through the optimization problem to both the predicted states of SAs and the cost weights\cite{amos2018differentiable}, thereby implicitly influencing the future behavior of SAs and adaptively adjusting the cost weights.

\subsubsection{Cost Function}
The cost function should account for important indicators such as velocity, acceleration changes, and turning effects to adapt to the requirements of complex traffic situations and improve operational efficiency and passenger comfort.

For the travel efficiency, to balance rapid destination arrival with traffic law compliance, we define the travel efficiency cost as the residual between the autonomous vehicle's instantaneous speed $V_t$ and the legal speed limit $V_\text{limit}$, as follows:

\begin{equation}
    j_{\text{speed}} = \sum_{t=1}^{T} (V_t-V_\text{limit}).
\end{equation}

For the passenger comfort, to enhance passenger comfort by ensuring smooth vehicle operation, we directly penalize the acceleration $a_t$, steering angle $\delta_t$, and their respective first-order time derivatives $\dot{a}_t$, $\dot{\delta}_t$. The corresponding cost function is designed as follows:
\begin{equation}
    \begin{aligned}
    j_{\text{acc}} =&  \sum_{t=1}^{T} a_t, \quad
    j_{\text{jerk}} =  \sum_{t=1}^{T} \dot{a}_t, \\
    j_{\text{steer}} =&  \sum_{t=1}^{T} \delta_t, \quad
    j_{\text{steer rate}} =  \sum_{t=1}^{T} \dot{\delta}_t .
\end{aligned}
\end{equation}

For the uncertainty-aware safety, it constitutes a critical component of our planning framework. Instead of enforcing safety with respect to a single predicted future, the proposed planner evaluates safety costs across multiple sampled future scenarios. This enables trajectory optimization to explicitly account for rare yet safety-critical future interactions. Since the diffusion module generates $M$ distinct possible states $\hat{S}=\{\hat{S}_{i,m}|m=1,2,...,M\},\hat{S}_{i}=\hat{S}_{1,...,N}^{1:T}$, each yielding substantially different SA trajectories, we incorporate chance constraints to robustly account for these $M$ probabilistic scenarios. We borrow the idea of calculating the safety cost in the Frenet frame from \cite{huang2023differentiable}, which allows for
effective tracking of the safe distance between the EV and the relevant agent. For the EV, the safety condition is satisfied if and only if $d_{\text{safe}}>\epsilon$. Here, the constant $\epsilon$  denotes the minimum safety clearance, defined as the aggregate of the two agents' lengths plus an additional safety margin. To reduce computational overhead, the safe distance is evaluated in the Frenet frame, considering solely the interacting agent pair, as follows:
\begin{equation}
    d_{\text{safe}} = \min \left \|p_t - \hat{p}_t^i  \right \| _2,
\end{equation}
where $\hat{p}_t^i$ is the predicted position of the interactive agent $i$ at future timestep $t$. For this problem, the multiple trajectories sampled from the learned conditional distribution of the diffusion model lack explicit probability estimates for individual trajectories. Therefore, our framework employs sampling to enforce chance constraints \cite{charnes1959chance}. Given that we generate $M$ trajectory samples, the original formulation can be transformed as follows:
\begin{equation}
    \frac{1}{M}\sum \mathbbm{1}(d_{\text{safe}} > \epsilon)\ge (1-\delta),
    \label{chance:1}
\end{equation}
where $\sum\mathbbm{1}(d_{\text{safe}} > \epsilon)$ represents the number of samples that satisfy the safety condition. Here, $\delta \in (0,1)$ denotes the risk tolerance parameter, representing the maximum acceptable probability of constraint violation. Equivalently, $(1-\delta)$ specifies the required confidence level for safety satisfaction. 

However, directly optimizing chance constraints is challenging in differentiable planning frameworks due to the non-smooth indicator function and the combinatorial nature of counting violations across sampled futures. Moreover, chance constraints only characterize violation frequency while ignoring the severity of unsafe outcomes, which is insufficient for safety-critical autonomous driving scenarios. To circumvent these issues, we transform the sample-based chance constraint into an empirical CVaR-based tail-risk constraint. The empirical CVaR-based tail-risk constraint serves as a tractable surrogate of the original chance constraint \cite{8767973}. We first transform (\ref{chance:1}) by replacing the proportion-based penalty with a distance-based formulation that penalizes the gap $d_{\text{gap}}$ between vehicles whenever safety constraints are violated. $d_{\text{gap}}$ is defined as follows:
\begin{equation}
d_{\text{gap}} =
\begin{cases}
     \epsilon-d_{\text{safe}}, &\epsilon\ge d_{\text{safe}} \\
     0, & \text{otherwise}.
\end{cases}
\end{equation}
For a random $d_{\text{gap}}$, the CVaR at confidence level $\delta$ is defined as the conditional expectation of losses exceeding the corresponding Value-at-Risk (VaR) threshold \cite{rockafellar2000optimization}:
\begin{equation}
\begin{aligned}
    & \text{VaR}_\delta(d_{\text{gap}}) = \inf\left\{ \alpha \in \mathbb{R} \,\middle|\, P(d_{\text{gap}} \leq \alpha) \geq 1-\delta \right\}, \\
    & \text{CVaR}_\delta(d_{\text{gap}}) = \mathbb{E}\left[ d_{\text{gap}} \,\middle|\, d_{\text{gap}} \geq \text{VaR}_\delta(d_{\text{gap}}) \right].
\end{aligned}
\end{equation}
When $\text{VaR}_\delta(d_{\text{gap}})=0$, (\ref{chance:1}) holds. Therefore, minimizing CVaR implicitly suppresses both the probability and severity of safety violations,
making it a conservative surrogate for the original chance constraint. Across the $M$ possible scenarios, at each timestep we sort the sampled violation magnitudes in descending order, yielding:
\begin{equation}
d_{\text{gap}}^1 \ge d_{\text{gap}}^2 \ge d_{\text{gap}}^3,..., \ge d_{\text{gap}}^M.
\end{equation}
We only penalize the mean of the $\left \lceil M\delta  \right \rceil $ largest $d_{\text{gap}}$, where $\left \lceil \right \rceil$ represents the ceil function. This corresponds to the risk tolerance $\delta$, focusing the penalty on the worst-case tail scenarios. Correspondingly, the empirical CVaR can then be approximated as:

\begin{equation}
    j_{\text{safety}} = \frac{1}{\left \lceil M\delta  \right \rceil}\sum_{t=1}^{T'}\sum_{i=1}^{\left \lceil M\delta  \right \rceil}d_{\text{gap}}^{i,t},
    \label{em:CVaR}
\end{equation}
 where $d_{\text{gap}}^{i,t}$ denotes the $i$-th order statistic in descending order. Equation (\ref{em:CVaR}) corresponds to the empirical tail expectation beyond the $\delta$-quantile and therefore serves as a discrete approximation of CVaR under the sampled future distribution. To enhance computational efficiency, we restrict safety cost evaluation to a sparse set of critical time instants $T' = [1,3,6,10,15,20,25,30,40,50]$.

The optimization problem is solved via the Gauss-Newton method. Since all aforementioned operations are differentiable, the predictor and planner can be jointly trained in an end-to-end manner, with the entire pipeline remaining fully differentiable. The planner requires an initial control sequence as an initial guess, which is provided by the predictor-generated control sequence $\hat{u}^{1:T}$.

\subsection{Learning Process}
\begin{figure}[t]
\includegraphics[width=1.0\linewidth]{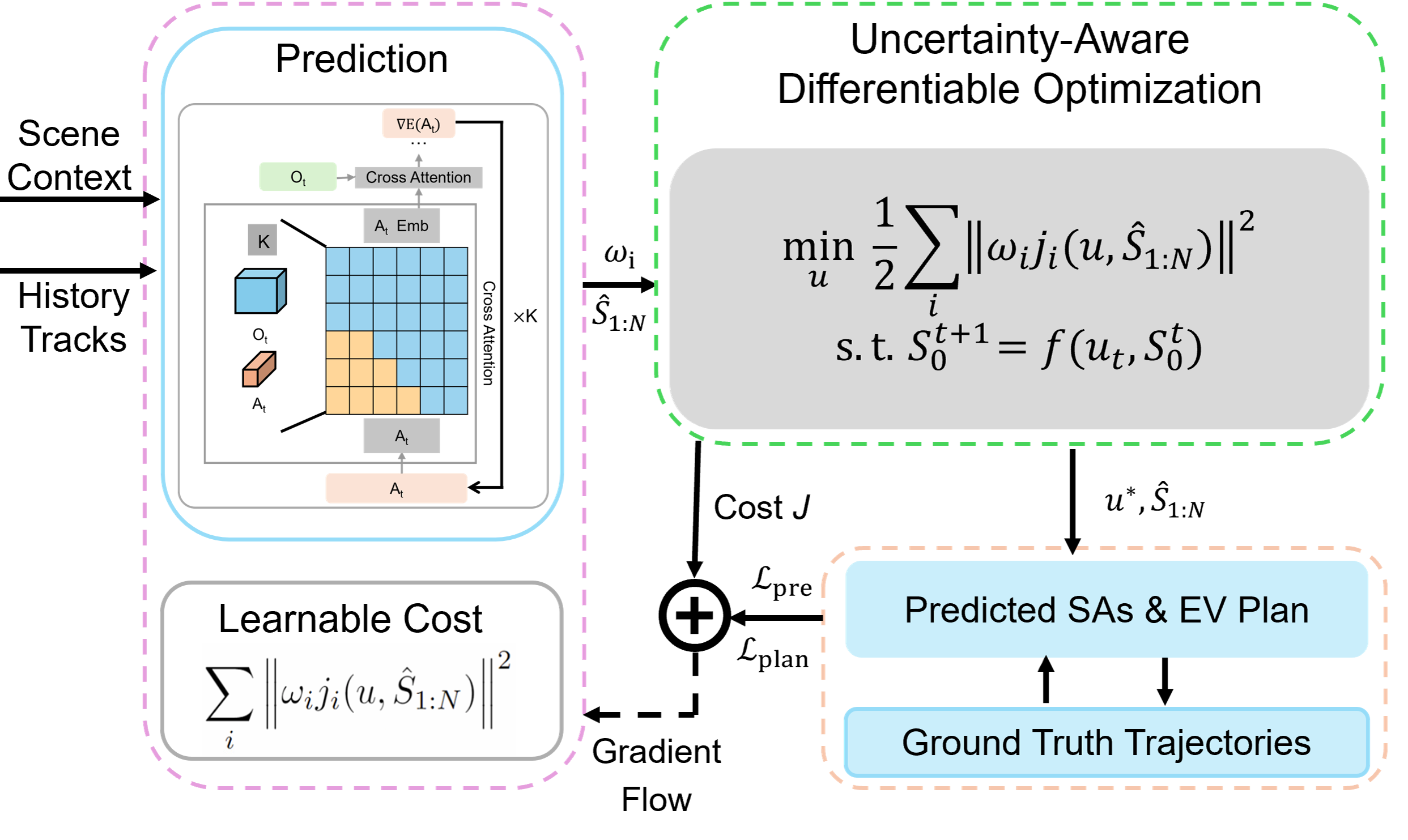}
   \caption{{Illustration of the joint training framework. During training, the diffusion predictor generates a single conditioned future realization together with an initial EV control sequence, which are fed into the differentiable planner. The planner optimizes the EV trajectory and computes planning losses against ground truth. Gradients are then backpropagated through the planner into the predicted trajectories and diffusion predictor, enabling end-to-end joint training. During inference, multiple future samples are generated for uncertainty-aware planning.}}
    \label{fig4}
\end{figure}

Overall, our framework comprises two stages, as illustrated in Fig.~\ref{fig4}. The first stage is the forward pass. The motion planner takes as inputs the predictions of SAs, an initial plan, and the cost function, then initiates trajectory optimization. The planning trajectory is iteratively refined until convergence criteria are satisfied or the maximum iteration count is reached. Upon optimization completion, the motion planner outputs a trajectory for the EV to execute.

Another stage is the backward pass. The loss function is evaluated on the optimized trajectory, and gradients are computed with respect to the initial plan, cost function weights, and SAs' predicted trajectories. These gradients are backpropagated through the neural network components to update parameters, thereby generating higher-quality trajectories in subsequent iterations. By integrating the differentiable optimization process with standard deep learning networks, we enhance the consistency between the learned parameters in network modules and the planning outcomes produced by the optimization process. The learning process of our framework is summarized in Algorithm \ref{alg1}. 

To make gradient propagation explicit, we adopt different prediction strategies during training and inference. During training, the diffusion predictor generates a single conditioned trajectory realization for SAs and an initial EV control sequence, which are fed into the differentiable planner. The planner optimizes the EV trajectory and computes planning losses, whose gradients are backpropagated through the optimization process to both the predicted SA trajectories and the predictor parameters. This establishes an end-to-end gradient pathway from the planning objective to the diffusion-based predictor. During inference, multiple trajectory samples are generated to represent predictive uncertainty and are jointly considered in the sample-conditioned planner for risk-aware optimization. Therefore, the distinction between training and inference lies only in the number of sampled futures, while the planner conditioning mechanism and optimization structure remain unchanged.

\begin{algorithm}[t]
\caption{Joint Training of Proposed Framework}
\label{alg1}
\begin{algorithmic}[1]
\REQUIRE Conditional diffusion-based predictor $\mathcal{D}_\theta$, sample-conditioned differentiable motion planner $\mathcal{P}$, and integrated objective $J$ of initial cost function weights $\omega=\{\omega_i\}$.

\STATE Construct the concatenated observation embedding by fusing scene context $\mathcal{M}$ with historical states $S_{0,1,...,N}^{-H:0}$.

\STATE Feed the concatenated observation embedding into the predictor $\mathcal{D}_\theta$, and output predicted trajectories $\hat{S}_{1:N}$ for SAs as well as an initial control sequence $\hat{u}$ for the EV:

\begin{center}
$\mathcal{D}_\theta(S_{0,1,...,N}^{-H:0},\mathcal{M})\to \hat{u},\hat{S}_{1:N}$
\end{center}

\STATE
Calculate prediction loss $\mathcal{L}_\text{prediction}$.

\STATE
Plan the EV’s trajectory $\hat{S}_{0}^{1:T}$ using the motion planner;

\begin{center}
$\mathcal{P}(\hat{S},\hat{u})\to \hat{S}_{0}^{1:T},\mathcal{L}_\text{cost}$. 
\end{center}

\STATE
Calculate planning loss $\mathcal{L}_\text{planning}$.
\STATE
Calculate total loss $\mathcal{L}$ according to (\ref{loss}).
\STATE
Backpropagate loss and calculate gradients with respect
to $\theta$ and $\omega$.
\STATE
Update $\mathcal{D}_\theta$ and $\omega$ using AdamW optimizer.
\end{algorithmic}
\end{algorithm}

For the joint training framework, the composite loss function consists of three primary terms: trajectory prediction loss for all SAs, trajectory planning loss for the EV, and the planning cost. The overall loss is defined as follows:
\begin{equation}
\label{loss}
    \mathcal{L}=\lambda_1 \mathcal{L}_\text{prediction}+\lambda_2 \mathcal{L}_\text{cost}+\lambda_3\mathcal{L}_\text{planning},
\end{equation}
where $\lambda_1$, $\lambda_2$, and $\lambda_3$ are the weights of the corresponding loss terms. For the prediction loss, we treat each future as a coherent joint future across all agents, backpropagating smooth L1 loss against ground-truth trajectories. During training, for computational efficiency and stable gradient propagation through the differentiable planner, the diffusion model generates a single conditioned trajectory realization from the learned conditional distribution. This sampled realization serves as the planner's input for computing planning gradients. During inference, multiple trajectory realizations are sampled to approximate the predictive distribution and enable uncertainty-aware planning. The prediction loss is therefore defined as follows:
\begin{equation}
\begin{split}
\mathcal{L}_{\text{prediction}} = 
\sum_{i=1}^N \biggl(
& \sum_{j=1}^T \text{smoothL}_1(\hat{S}_{i,j} - S_{i,j}^{\text{gt}}) \\
& + \text{smoothL}_1(\hat{S}_{i,T} - S_{i,T}^{\text{gt}})
\biggr),
\end{split}
\end{equation}
where $S_i^{gt}$ is the individual ground truth trajectory and $T$ is the total future timesteps. We compute the smooth L1 loss at the final timestep twice to mitigate endpoint drift.

The motion planner takes the predicted trajectories of SAs and the initial control sequence of the EV as inputs, where the former are utilized for safety cost computation. The motion planner outputs the planned EV trajectory $\hat{S}_0$, which is subsequently compared against the ground-truth EV trajectory $S_0^{gt}$ to evaluate the planning loss $\mathcal{L}_\text{planning}$:
\begin{equation}
\begin{split}
    \mathcal{L}_\text{planning}= & \sum_{i=1}^T \text{smoothL}_1(\hat{S}_{0,i}-S_{0,i}^{gt}) \\
    & +\text{smoothL}_1(\hat{S}_{0,T}-S_{0,T}^{gt}).
\end{split}
\end{equation}
$\mathcal{L}_\text{cost}$ denotes the planning cost function value, which is also output by the motion planner.

\section{Experiment Results}\label{sec:exp}
\subsection{Experiment Setup}
\subsubsection{Testing Benchmarks}
To evaluate our framework's performance in prediction and planning, we conduct model training and testing on two distinct datasets. This dual-dataset validation serves to verify the efficacy of the proposed approach. The two datasets are described as follows:
\begin{enumerate}
    \item Waymo Open Motion Dataset (WOMD) \cite{ettinger2021large}: This large-scale real-world driving dataset centers on urban scenarios featuring intricate road layouts and varied traffic interactions. contains 103,354 distinct 
    20$\,$s driving episodes paired with fine-grained map information and agent trajectories. We randomly selected 10\% of the dataset, i.e., 100 data files from a total of 1,000 files. Random storage of scenes guarantees the subset maintains the original data distribution. A 7$\,$s temporal window is adopted to split each 20$\,$s scenario into frames at 10$\,$Hz, covering 2$\,$s historical observations and 5$\,$s prediction horizon. The window slides forward with a 10-timestep (1$\,$s) stride, generating 14 segmented samples per episode. The sampled data is split into 80\% training set and 20\% validation \& open-loop test set, containing 73005 and 18095 frames, respectively.
    \item Argoverse 2 Motion Forecasting Dataset \cite{wilsonargoverse}:
    The motion prediction dataset comprises 250,000 scenarios capturing interactions between the autonomous vehicle and other agents within local contexts. The task requires predicting the future motion of scored actors, which are high-quality trajectories of SAs in proximity to the EV. We randomly sample 22,500 scenarios and apply identical preprocessing as described for the WOMD. This yields a total of 70,000 frames for training, with 20,000 frames allocated for validation and open-loop evaluation.
\end{enumerate}

\subsubsection{Testing Metrics}
The framework is validated under both open-loop and closed-loop conditions. During open-loop tests, the motion planner synthesizes a $5\,$s EV trajectory using the present state and $5\,$s SA predictions, which is subsequently benchmarked against the actual EV trajectory. For closed-loop evaluation, we develop a simulator in which the EV implements the first action of its planned trajectory at each step to refresh its state, while SAs move along their dataset-recorded paths. The following metrics are adopted for performance quantification.
\begin{enumerate}
    \item Safety: For safety evaluation, we primarily consider collision rate and off-route rate \cite{huang2023differentiable}. We perform collision analysis between the planned EV trajectory and ground-truth trajectories of SAs, and compute the off-route rate by comparing the planned EV trajectory against reference lane boundaries. In open-loop testing, a collision or off-route event at any planning timestep is counted as a collision or off-route occurrence. In closed-loop testing, we perform collision and off-route detection at every simulation step, and we adopt the success rate as the evaluation metric, which indicates no collisions or off-route deviations.
    \item Planning: For planning accuracy, we primarily compare position errors at different timesteps. For open-loop testing, we evaluate position errors at 1$\,$s, 3$\,$s, and 5$\,$s. For closed-loop testing, we compare errors at 3$\,$s, 5$\,$s, and 10$\,$s on the WOMD. Due to the shorter total sequence length in the Argoverse 2 motion forecasting dataset, we compare errors at 3$\,$s, 5$\,$s, and 8$\,$s.
    \item Vehicle Dynamics: We introduce three metrics to quantify comfort and dynamic feasibility: longitudinal acceleration, longitudinal jerk, and lateral acceleration. These are computed as time-averaged absolute values per scenario and benchmarked against human driving statistics.
    \item Motion Prediction: For motion prediction, we primarily consider accuracy guarantees under diffusion-based trajectory sampling. Therefore, we employ minimum average displacement error (minADE) and minimum final displacement error (minFDE) to reflect prediction accuracy. The  minADE and minFDE are defined as follows:
    \begin{align}
    \text{minADE} &= \min_M\Bigg( \frac{1}{T \times N} \nonumber \\
    & \qquad \times \sum_i^T\sum_j^N\left\| (\hat{x},\hat{y})_j^i-(x,y)_j^{i,\text{gt}} \right\|_2 \Bigg), \\
    \text{minFDE} &= \min_M\Bigg( \frac{1}{N}\sum_j^N\left\| (\hat{x},\hat{y})_j^T-(x,y)_j^{T,\text{gt}} \right\|_2 \Bigg).
    \end{align}
    where $T$ denotes the maximum timestep, $N$ represents the number of agents, $(\hat{x},\hat{y})$ indicates the predicted coordinates, and $(x,y)^{gt}$ represents the ground truth coordinates.
    
\end{enumerate}

\subsubsection{Comparison Baselines}
We compare the proposed framework against the following baselines to demonstrate the advantages and effectiveness of our approach.
\begin{enumerate}
    \item Vanilla IL: Based on identical scene contexts, we train a network using pure IL to directly generate the EV trajectory. The network architecture mirrors our proposed framework, with SAs' predictions removed, which means outputting a single ego trajectory rather than sampling $M$ trajectories from the learned conditional distribution. The IL policy is trained on the same dataset with identical hyperparameters as our framework, ensuring a fair comparison.
    \item IL with Prediction Subtask: Similar to the IL baseline above, we train a network using IL to directly generate trajectories for both the EV and SAs. The network architecture remains identical to our proposed framework, sampling $M$ trajectories from the learned conditional distribution. We select the trajectory with the minimum prediction error as the EV's final trajectory. This policy is trained on the same dataset with identical hyperparameters as our framework, ensuring a fair comparison.
    \item Differentiable Integrated Prediction and Planning (DIPP) \cite{huang2023differentiable}: We utilize DIPP, a well-known learning-based framework that also adopts a differentiable integrated prediction and planning framework. DIPP employs a structured learning architecture that couples motion prediction with trajectory planning, leveraging pre-configured reference routes. However, it considers only the single best prediction during planning. By comparing our framework against this baseline, we aim to validate the effectiveness of our approach and underscore the importance of accounting for predictive uncertainty in the planning process.
    \item GameFormer \cite{Huang_2023_ICCV}: We adopt GameFormer as a comparative baseline, which is an interactive prediction and planning framework grounded in hierarchical game theory. GameFormer employs a Transformer encoder-decoder architecture with a hierarchical interaction decoder that iteratively refines future trajectories for all agents. This framework has achieved state-of-the-art performance on both the Waymo interaction prediction task and the nuPlan planning benchmark. By comparing our framework against GameFormer, we aim to validate the effectiveness of our approach in interactive prediction and joint planning, while highlighting the advantages of differentiable optimization in the planning process.

\end{enumerate}

\subsubsection{Implementation Details}
The model generates future scenarios in the form of joint trajectories encompassing all agents. For SAs, we regress relative displacements from their instantaneous locations instead of absolute world coordinates, yielding substantial gains in predictive accuracy. For the EV, the forecasted trajectory is fed into an MLP to produce a control sequence, which is adopted as the initial control input for the motion planner. Furthermore, the network embeds the learning of cost function weights, which is realized by a lightweight MLP fed with fixed dummy inputs to generate the weight parameters. The collision cost weight is fixed to a sufficiently large constant and frozen during training to enforce constraint satisfaction.

For the training strategy, we first pre-train the diffusion model independently until its outputs stabilize, before proceeding to end-to-end joint training. During training, the diffusion model generates samples via single-step diffusion. Considering computational efficiency, we cap the motion planner at a maximum of 2 iterations, a design choice that further encourages the network to output high-quality trajectories. The step size for the Gauss-Newton update is $\alpha=0.4$. We apply gradient clipping to the network parameters with a maximum norm of 5. 
During testing, the diffusion model simultaneously generates 10 samples, reflecting the uncertainty in vehicle motion. To ensure high-quality trajectory generation for the EV, the motion planner is configured with an iteration cap of 50, a step size of 0.2, and an absolute error tolerance of $10^{-2}$.

The primary model configurations and training settings utilized in our experiments are outlined in Table \ref{tab:1}. The parameters used for training and testing on the two datasets are the same.

\begin{table}[t]
  \centering
  \caption{hyperparameters of the main modules during training and inference process}
  \resizebox{0.75\linewidth}{!}{
  \begin{footnotesize}
    \begin{tabular}{cc}
    \toprule
    \multicolumn{1}{c}{Parameter} & \multicolumn{1}{c}{Value} \\
    \midrule
    Number of neighbors $N$ & 10 \\
    Number of predicted future $M$ & 10 \\
    Past timesteps $H$ & 20 \\
    Future timesteps $T$ &  50 \\
    \midrule
    Optimizer  & Adam \\
    Scheduler  & Cosine \\
    Warm-up steps  & 2000 \\
    Initial learning rate   & 5.00e-04 \\
    \midrule
    \textbf{Training} \\
    Step size $\alpha$ & 0.4 \\
    Planning iterations  & 2 \\
    Total training epochs & 100 \\
    Pretraining epochs    & 80 \\
    Embedding size & 256 \\
    Batch size    & 512 \\
    Diffusion iterations (DDIM \cite{songdenoising})  & 100 \\
    Weight for prediction loss $\lambda_1$ & 1 \\
    Weight for planning cost  $\lambda_2$ & 1 \\
    Weight for planning loss $\lambda_3$ & 1 \\
    \midrule
    \textbf{Testing} \\
    Step size $\alpha$ & 0.2 \\
    Planning iterations  & 50 \\
    Risk tolerance $\delta$ & 0.1 \\
    Diffusion iterations (DDIM \cite{songdenoising})  & 20 \\
    \bottomrule
    \end{tabular}%
    \end{footnotesize}
    }\label{tab:1}%
\end{table}%

\subsection{Open-Loop Testing Results}
In open-loop testing, the EV predicts SAs' trajectories over a 5$\,$s horizon and plans its own 5$\,$s trajectory accordingly. We evaluate performance by comparing the EV's planned 5$\,$s trajectory, the predicted 5$\,$s trajectories of SAs, and the ground-truth trajectories of all vehicles. The quantitative open-loop comparison between our proposed framework and baselines is presented in Table \ref{tab:3}.

As evidenced by the experimental results, our approach outperforms baselines in terms of both minADE and minFDE. The second-best performance on these two metrics is achieved by the IL with the prediction subtask baseline, which employs an identical prediction network structure. For this reason, we focus our comparison on DIPP, the third-best framework. Relative to DIPP, our framework boosts minADE by 20.99\% and 51.54\%, and minFDE by 74.09\% and 87.74\% on the WOMD and Argoverse 2 datasets, respectively.
Typically, FDE exceeds ADE due to cumulative errors over time. In our framework, however, the introduction of an extra terminal loss alleviates endpoint error. Meanwhile, the denoising process of the diffusion model, which is characterized by autoregressive or holistic generation, prioritizes the stability of the trajectory tail, causing FDE to become smaller than ADE. We logged a subset of prediction errors during testing and confirmed that most errors peak in the middle segment and decrease toward both ends. This pattern is further supported by the planning error of Vanilla IL, whose trajectories are directly output by the prediction network and similarly show larger errors at 3$\,$s than at 5$\,$s.

In addition to prediction performance, our framework also achieves remarkable results in motion planning. It attains the minimum collision rate and off-route rate across both datasets, corresponding to reductions of (74.12\%, 32.11\%) for collision rate and (71.15\%, 3.21\%) for off-route rate relative to the suboptimal counterparts, respectively. It also ranks first or second in all vehicle dynamic metrics, verifying that our approach guarantees reliable driving comfort. Regarding planning accuracy, Vanilla IL and IL with prediction subtask exhibit artificially low errors since planning trajectories are directly regressed from the prediction network without physical constraints. Nevertheless, vehicle metrics indicate that these small errors are unrealistic and non-physical, thus lacking practical value in real driving scenarios. By comparison, the remaining frameworks preserve physically consistent motion and reasonable vehicle behaviors. Within this group, our framework outperforms baselines in terms of planning error. The safety improvement mainly stems from conditioning trajectory optimization on multiple sampled future scenarios, which allows the planner to account for uncertain interactions during optimization rather than relying on a single deterministic prediction.

\begin{figure*}[htbp]
\centering
\setlength{\fboxsep}{2pt}    
\setlength{\fboxrule}{1pt} 

\subfloat{\fbox{\includegraphics[width=0.18\linewidth,height=3.55cm]{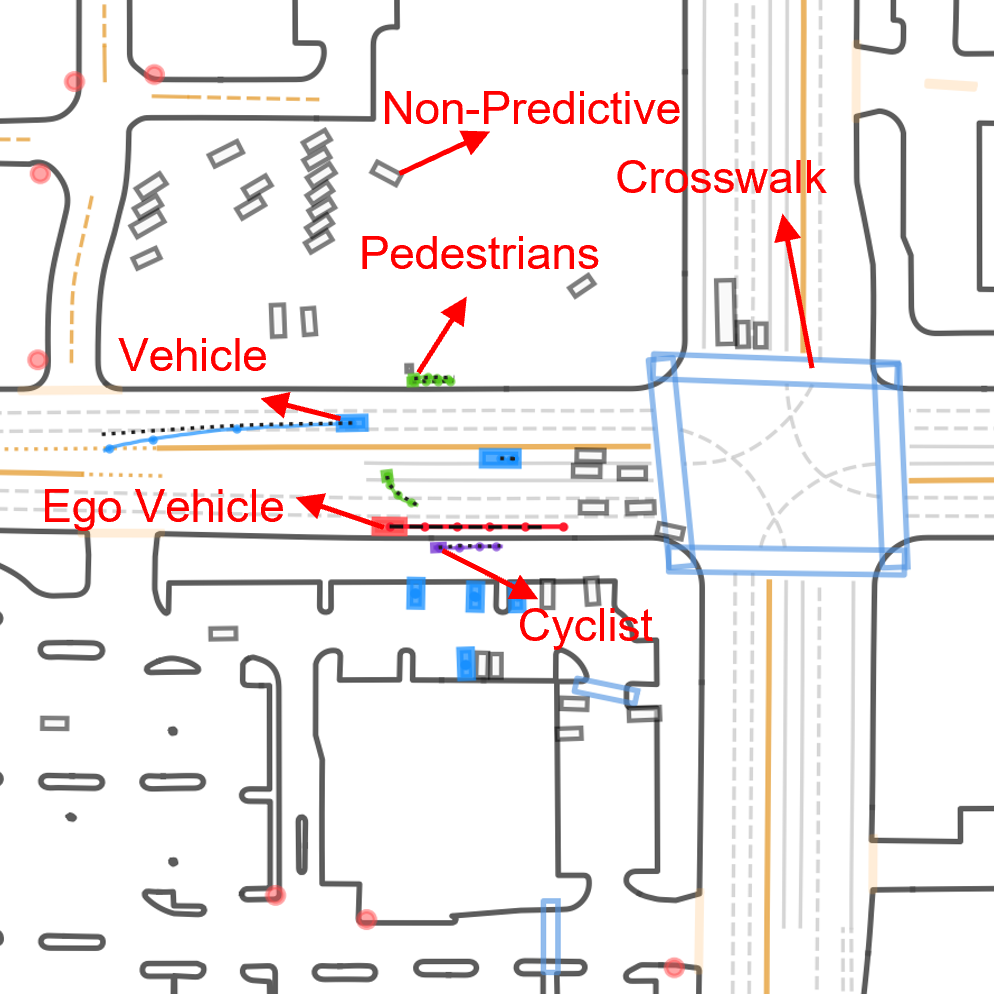}}} \hspace{0.5mm}
\subfloat{\fbox{\includegraphics[width=0.18\linewidth,height=3.55cm]{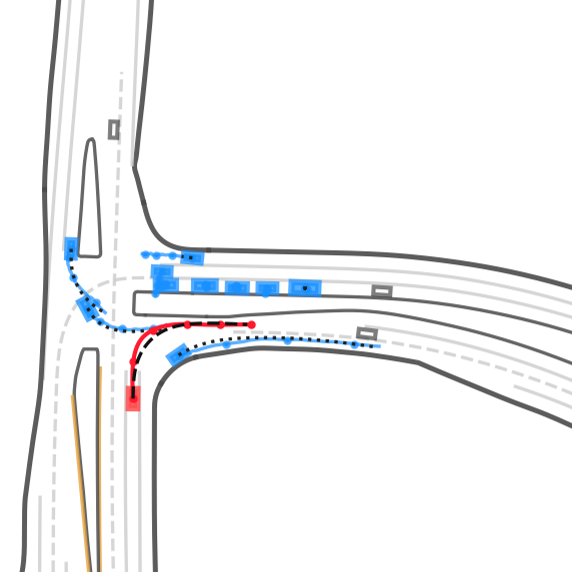}}} \hspace{0.5mm}
\subfloat{\fbox{\includegraphics[width=0.18\linewidth,height=3.55cm]{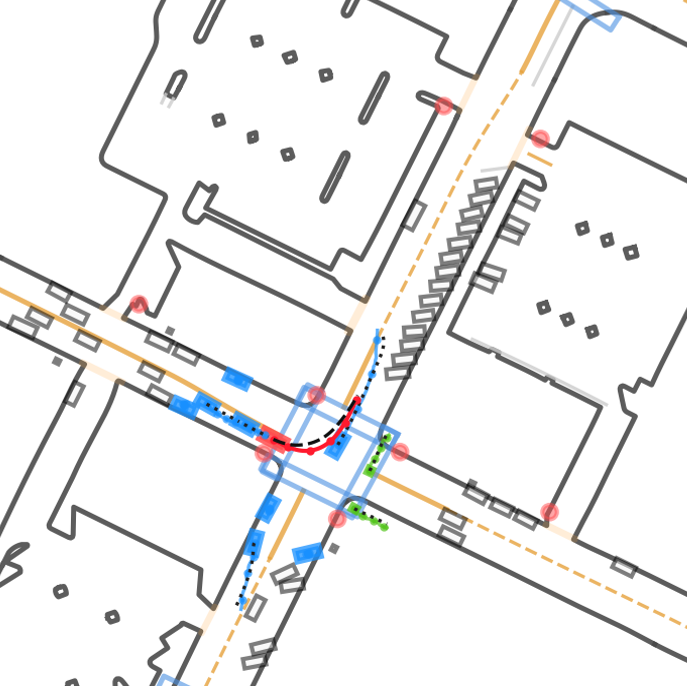}}} \hspace{0.5mm}
\subfloat{\fbox{\includegraphics[width=0.18\linewidth,height=3.55cm]{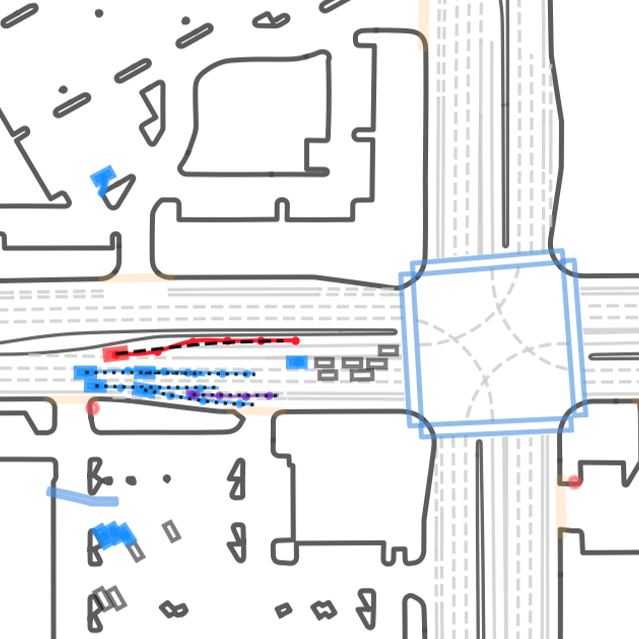}}} \hspace{0.5mm}
\subfloat{\fbox{\includegraphics[width=0.18\linewidth,height=3.55cm]{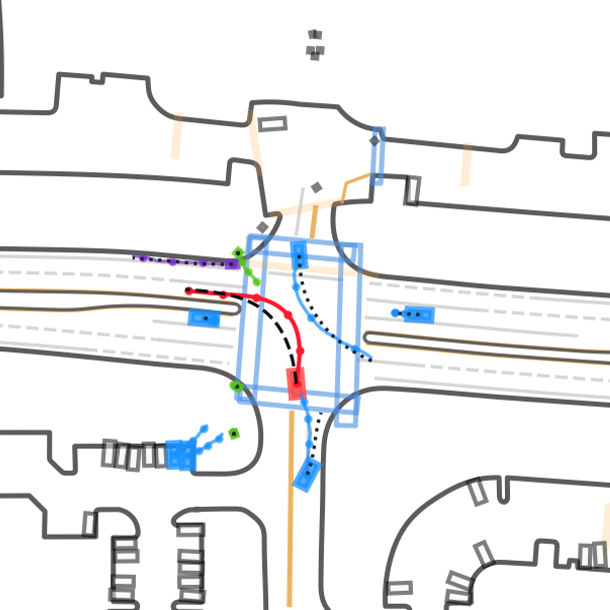}}} \\[-1mm]

\subfloat{\fbox{\includegraphics[width=0.18\linewidth,height=3.55cm]{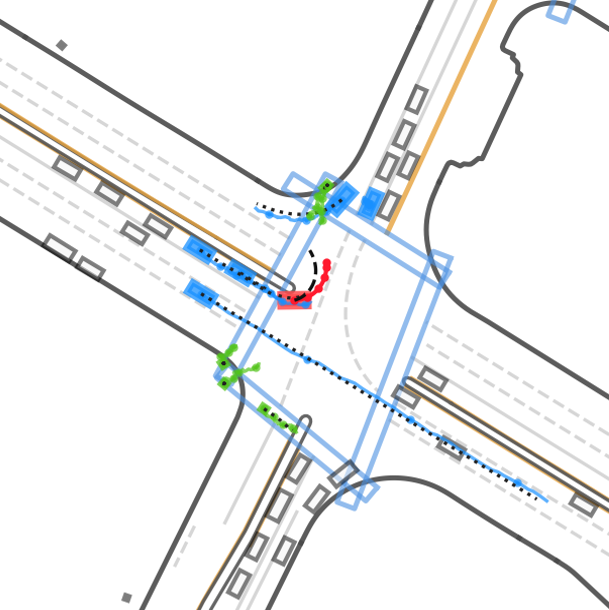}}} \hspace{0.5mm}
\subfloat{\fbox{\includegraphics[width=0.18\linewidth,height=3.55cm]{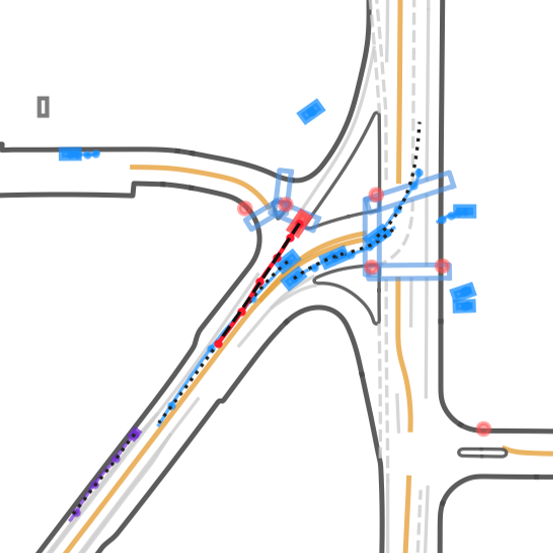}}} \hspace{0.5mm}
\subfloat{\fbox{\includegraphics[width=0.18\linewidth,height=3.55cm]{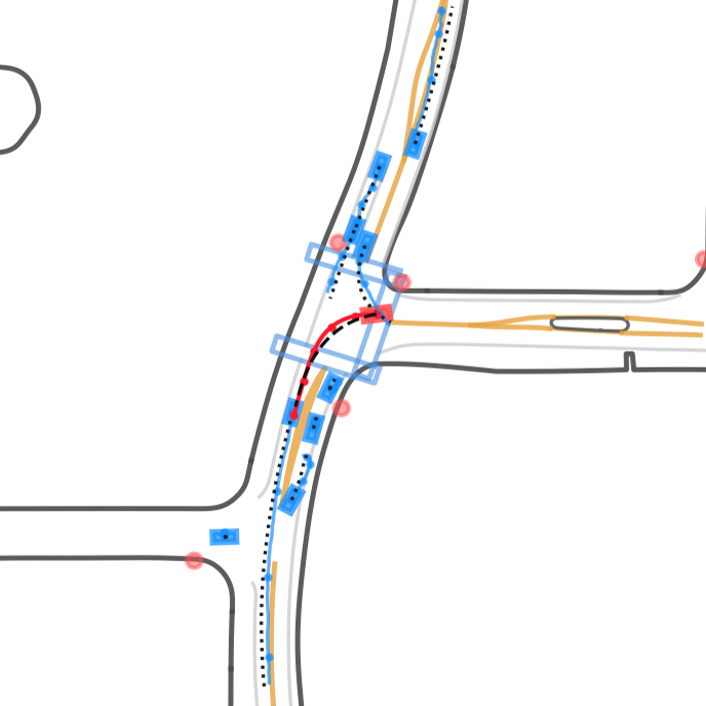}}} \hspace{0.5mm}
\subfloat{\fbox{\includegraphics[width=0.18\linewidth,height=3.55cm]{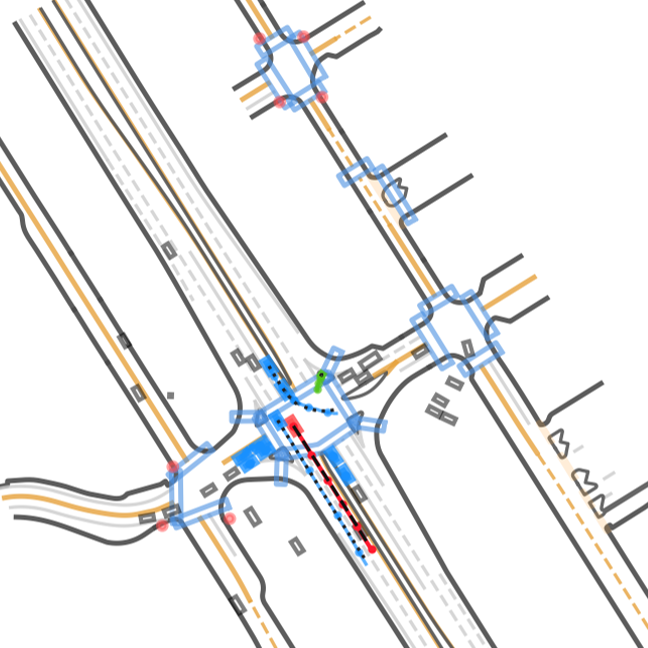}}} \hspace{0.5mm}
\subfloat{\fbox{\includegraphics[width=0.18\linewidth,height=3.55cm]{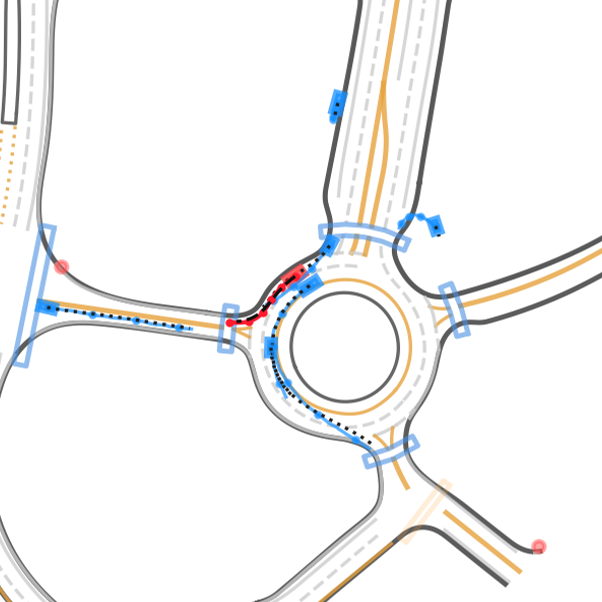}}} \\[-1mm]

\subfloat{\fbox{\includegraphics[width=0.18\linewidth,height=3.55cm]{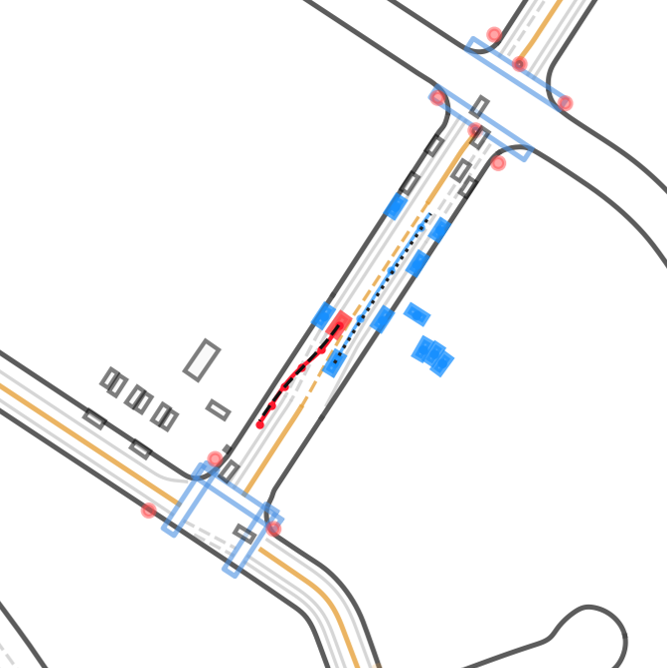}}} \hspace{0.5mm}
\subfloat{\fbox{\includegraphics[width=0.18\linewidth,height=3.55cm]{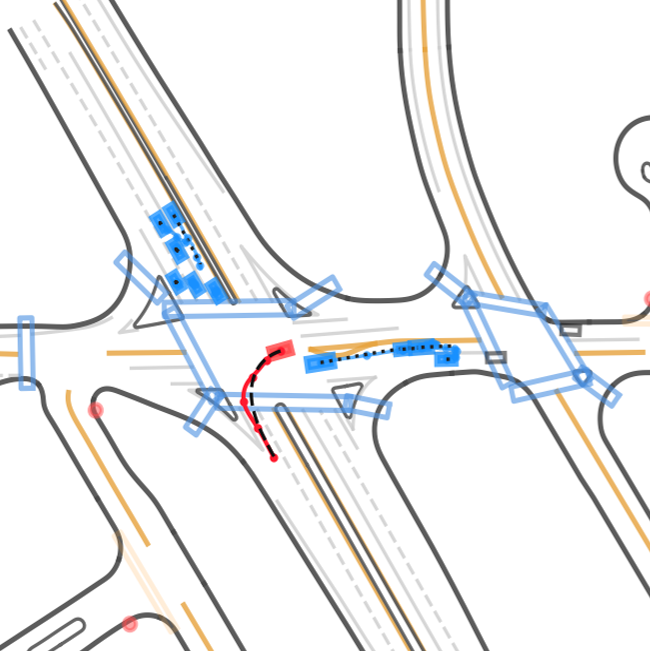}}} \hspace{0.5mm}
\subfloat{\fbox{\includegraphics[width=0.18\linewidth,height=3.55cm]{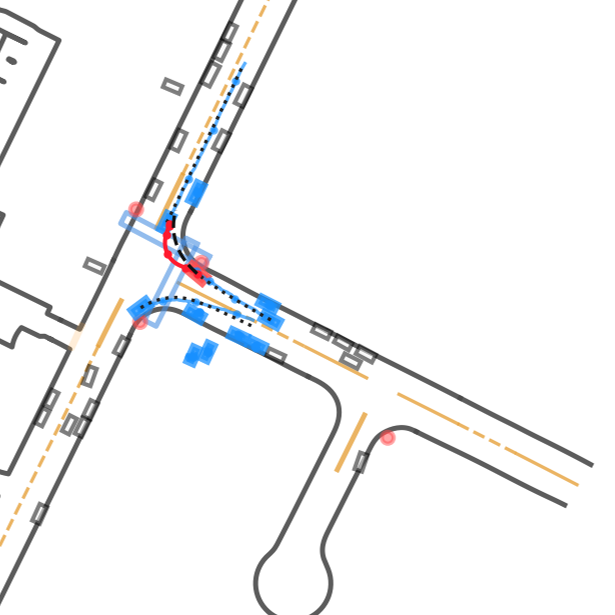}}} \hspace{0.5mm}
\subfloat{\fbox{\includegraphics[width=0.18\linewidth,height=3.55cm]{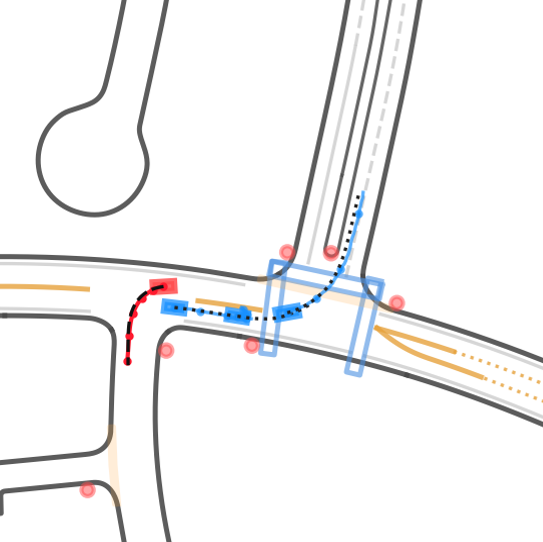}}} \hspace{0.5mm}
\subfloat{\fbox{\includegraphics[width=0.18\linewidth,height=3.55cm]{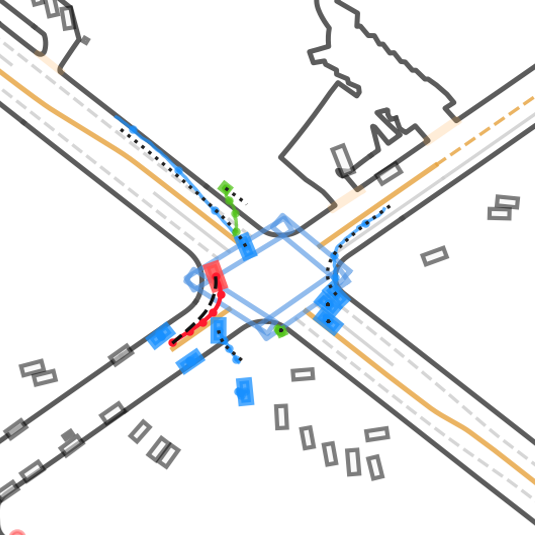}}} \\[-1mm]

\subfloat{\fbox{\includegraphics[width=0.18\linewidth,height=3.55cm]{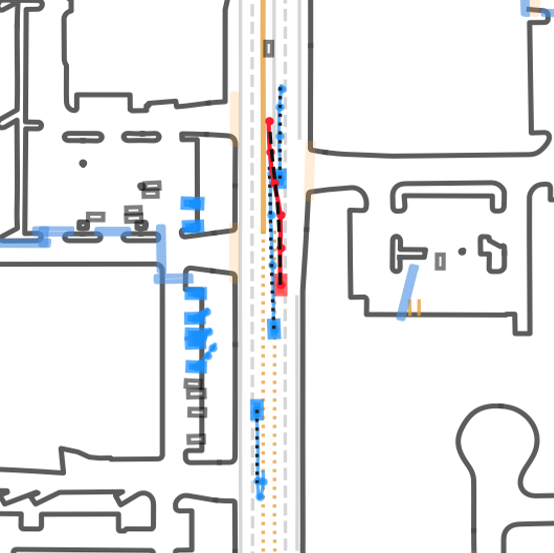}}} \hspace{0.5mm}
\subfloat{\fbox{\includegraphics[width=0.18\linewidth,height=3.55cm]{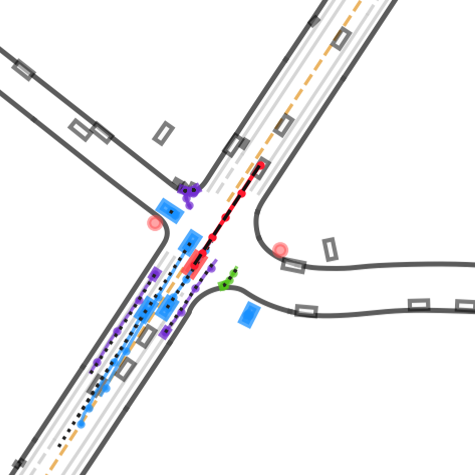}}} \hspace{0.5mm}
\subfloat{\fbox{\includegraphics[width=0.18\linewidth,height=3.55cm]{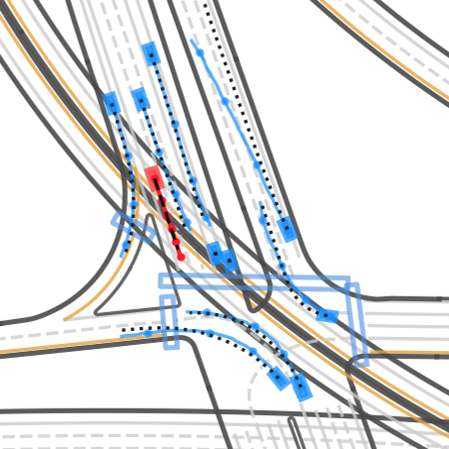}}} \hspace{0.5mm}
\subfloat{\fbox{\includegraphics[width=0.18\linewidth,height=3.55cm]{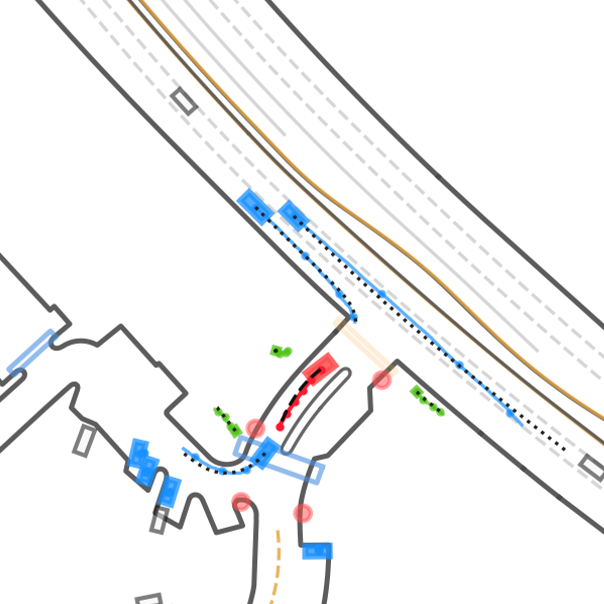}}} \hspace{0.5mm}
\subfloat{\fbox{\includegraphics[width=0.18\linewidth,height=3.55cm]{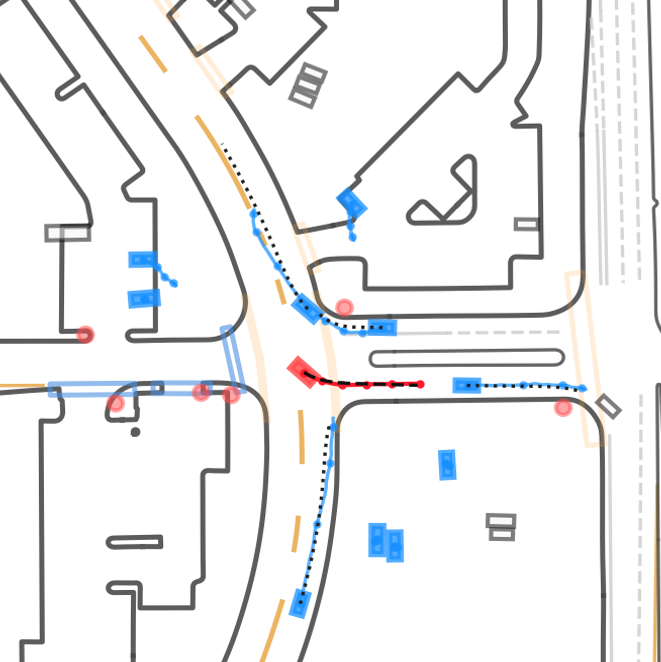}}} \\[-1mm]

 \caption{Qualitative evaluation of the proposed framework across diverse open-loop scenarios. Colored solid lines represent the planned trajectory for the EV and the predicted trajectories for SAs, while black dashed lines denote the corresponding ground truth. For visual clarity, only the best-matching predicted sample from the diffusion model is displayed for each SA.
   }
    \label{Fig5}
\end{figure*}

We also conduct qualitative experiments. Fig.~\ref{Fig5} illustrates the results in several representative scenarios. It can be observed that our framework provides accurate trajectory predictions for SAs while generating a safe trajectory for the EV. Moreover, our framework applies to common urban scenarios, including straight roads, intersections, ramps, and roundabouts. Compared to the ground-truth trajectories, our planner produces trajectories with larger turning radii and smoother curvature in cornering scenarios, as it accounts for driving comfort.

 \begin{table*}[t]
          \centering
          \caption{comparison of the proposed framework with baselines in open-loop testing.}
          \renewcommand{\arraystretch}{1.1}
          \subfloat[WOMD Scenario]{
            \begin{tabular}{c|cc|ccc|ccc|cc @{\extracolsep{\fill}}}
            \toprule
            \multirow{2}{*}{Method} & Collision $\downarrow$ & Off-route $\downarrow$ & Acc. $\downarrow$ & Jerk $\downarrow$ & Lat. Acc. $\downarrow$ & \multicolumn{3}{c|}{Planning error (m) $\downarrow$} & \multicolumn{2}{c}{Prediction error (m) $\downarrow$} \\
                  & (\%)  & (Thre.=5$\,$m)(\%)   & (m/s$^2$) & (m/s$^3$) & (m/s$^2$) & @1$\,$s   & @3$\,$s   & @5$\,$s & minADE  & minFDE
                   \\
            \midrule
            Vanilla IL & 30.60 & 5.89 & 9.02 & 61.20 & \underline{8.66} & \textbf{0.22} & \textbf{0.30} & \textbf{0.25} & $\slash$ & $\slash$  \\
            IL + Prediction & 30.19 & 5.32 & 18.96 & 122.59 & 18.60 & 0.54 & 0.80 & \underline{0.60} & \underline{0.67} & \underline{0.53} \\
            DIPP \cite{huang2023differentiable} & \underline{3.71} & \underline{3.05} & \underline{0.49} & 0.74 & \textbf{0.00} & 0.29 & 1.18 & 3.03 & 0.81 & 1.93 \\
            GameFormer \cite{Huang_2023_ICCV} & 6.87 & 9.31 & \textbf{0.47} & \underline{0.70} & \textbf{0.00} 
            & 0.77 & \underline{0.77} & 2.36 & 0.87 & 2.01 \\
            Ours & \textbf{0.96} & \textbf{0.88} & 0.52 & \textbf{0.69} & \textbf{0.00} & \underline{0.27} & 0.79 & 1.20 
            & \textbf{0.64} & \textbf{0.50}  \\
            \midrule
            Human & $\slash$ & $\slash$ & $0.67$ & $3.41$ & $0.10$ & $\slash$ & $\slash$ & $\slash$ & $\slash$ & $\slash$  \\
            \bottomrule
            \end{tabular}%
            \label{tab:31}%
          }
        \hspace{1mm}
        \subfloat[Argoverse 2 Motion Forecasting Dataset Scenario]{

            \begin{tabular}{c|cc|ccc|ccc|cc @{\extracolsep{\fill}}}
            \toprule
            \multirow{2}{*}{Method} & Collision $\downarrow$ & Off-route $\downarrow$ & Acc. $\downarrow$ & Jerk $\downarrow$ & Lat. Acc. $\downarrow$ & \multicolumn{3}{c|}{Planning error (m) $\downarrow$} &\multicolumn{2}{c}{Prediction error (m) $\downarrow$} \\
                  & (\%)  & (Thre.=5$\,$m)(\%)  & (m/s$^2$) & (m/s$^3$) & (m/s$^2$) & @1$\,$s   & @3$\,$s   & @5$\,$s & minADE  & minFDE
                   \\
            \midrule
            Vanilla IL & 14.15 & 9.85 & 3.11 & 46.95 & \underline{3.59} & \textbf{0.24} & \textbf{0.31} & \textbf{0.20} & $\slash$ & $\slash$ \\
            IL + Prediction & 14.41 & 10.26 & 16.05 & 93.05 & 15.67 & \underline{0.78} & \underline{1.08} & \underline{0.57} & \underline{0.68} & \underline{0.42}  \\
            DIPP \cite{huang2023differentiable} & 12.37 &  \underline{5.61} & 0.66 & 1.34 & \textbf{0.00} & 1.11 & 2.64 & 5.32 & 1.30 & 3.10 \\
            GameFormer\cite{Huang_2023_ICCV} & \underline{6.82} & 7.00 &  \underline{0.60} &  \underline{1.22} & \textbf{0.00} & 0.95 & 1.78 & 3.44 & 1.94 & 3.78  \\
            Ours & \textbf{4.63} & \textbf{5.43} & \textbf{0.58} & \textbf{0.58} & \textbf{0.00} & 1.09 & 1.13 & 3.15 & \textbf{0.63} & \textbf{0.38} \\
            \midrule
            Human & $\slash$ & $\slash$ & 0.54 & 0.84 & 0.02 & $\slash$ & $\slash$ & $\slash$ & $\slash$ & $\slash$  \\
            \bottomrule
            \end{tabular}%
            \label{tab:32}%
          }
        \hspace{1mm}
        \begin{minipage}{0.95\linewidth}
        \footnotesize
        \vspace{0.5em}
        ACC.: acceleration, Lat. ACC.: lateral acceleration, Thre.: threshold. \textbf{x.x} denotes the optimal result, and \underline{x.x} denotes the suboptimal result. All data are presented to two decimal places. $\downarrow$ denotes that lower values signify better results. $\uparrow$ denotes that higher values signify better results.
        \end{minipage}
        \label{tab:3}%
        \end{table*}%

\subsection{Closed-Loop Testing Results}

In closed-loop testing, we randomly selected 100 scenarios to evaluate the effectiveness of our proposed framework. To assess closed-loop performance, we introduce an additional progress metric, which measures the travel distance of the autonomous vehicle before reaching the scenario endpoint, colliding with other agents, or deviating from the route. The closed-loop testing results are presented in Table \ref{tab:4}.

\begin{figure*}[htbp]
\centering
\setlength{\fboxsep}{2pt}    
\setlength{\fboxrule}{1pt} 

\subfloat{\fbox{\includegraphics[width=0.18\linewidth,height=3.55cm]{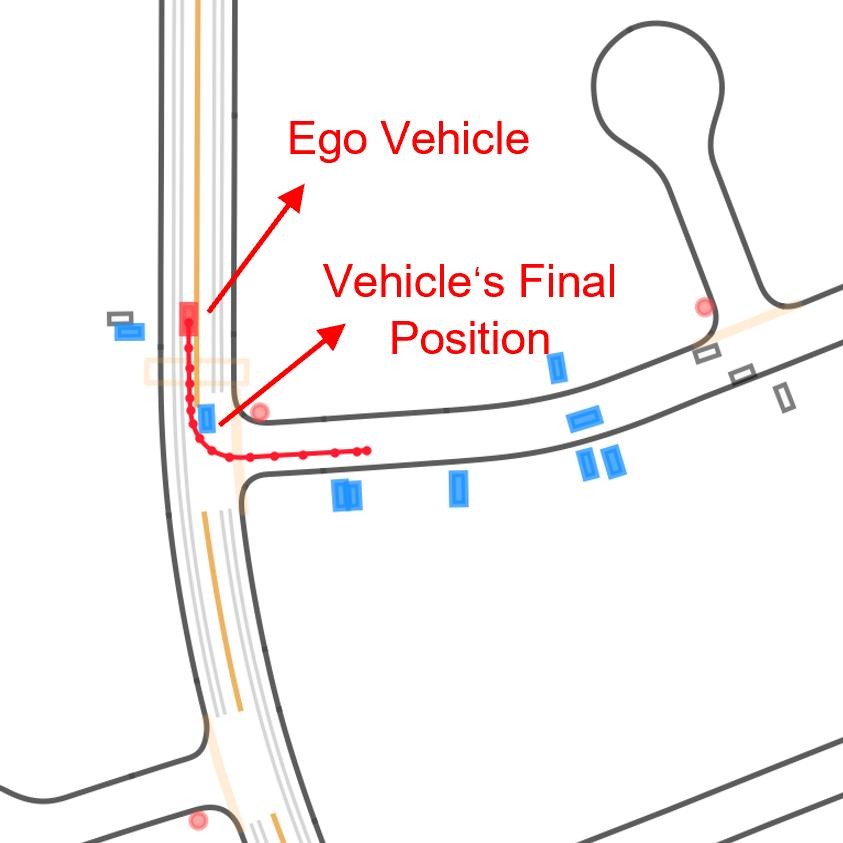}}} \hspace{0.5mm}
\subfloat{\fbox{\includegraphics[width=0.18\linewidth,height=3.55cm]{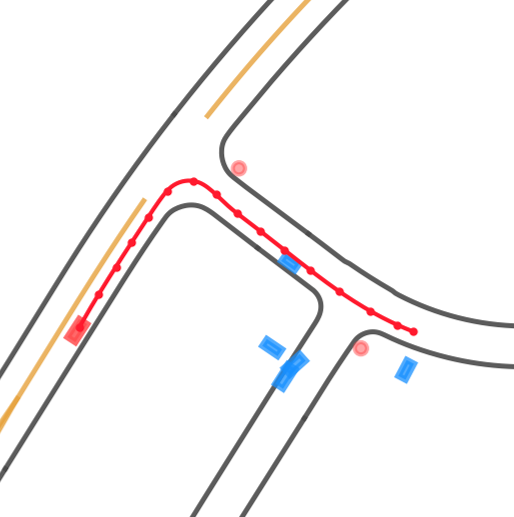}}} \hspace{0.5mm}
\subfloat{\fbox{\includegraphics[width=0.18\linewidth,height=3.55cm]{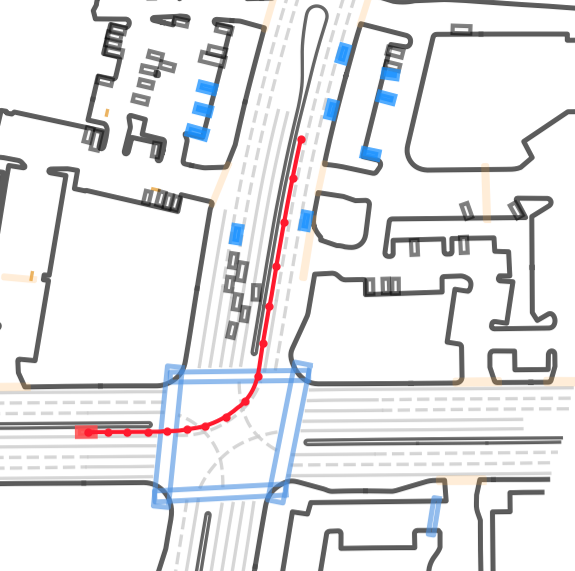}}} \hspace{0.5mm}
\subfloat{\fbox{\includegraphics[width=0.18\linewidth,height=3.55cm]{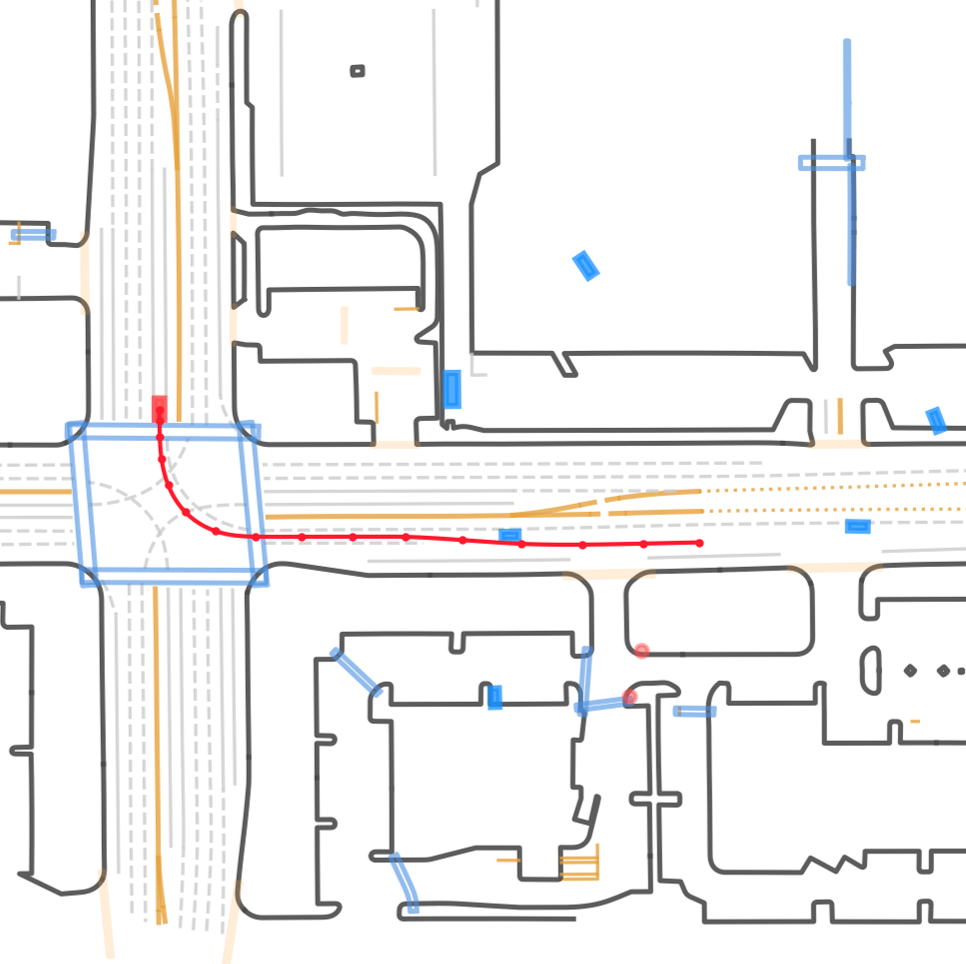}}} \hspace{0.5mm}
\subfloat{\fbox{\includegraphics[width=0.18\linewidth,height=3.55cm]{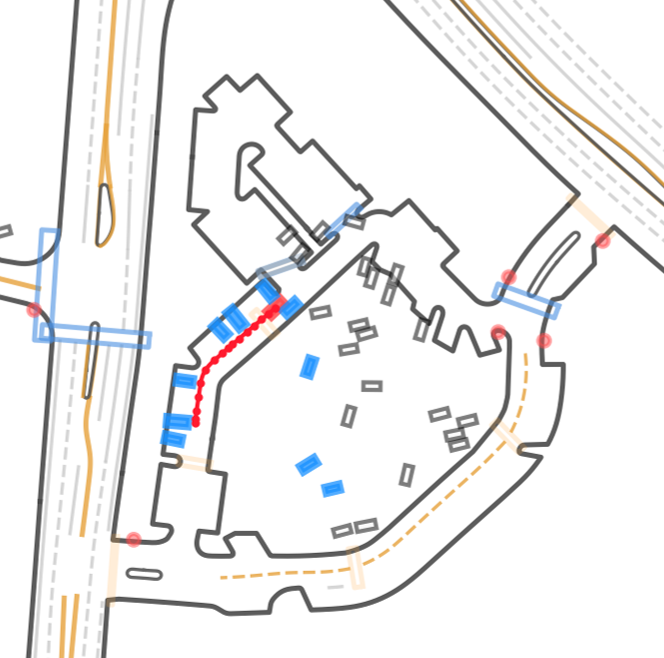}}} \\[-1mm]

\subfloat{\fbox{\includegraphics[width=0.18\linewidth,height=3.55cm]{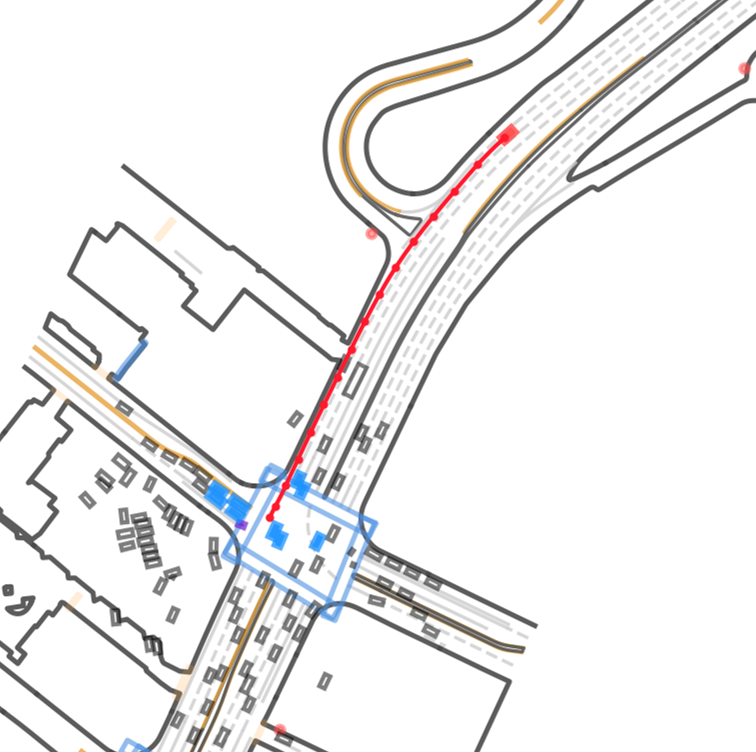}}} \hspace{0.5mm}
\subfloat{\fbox{\includegraphics[width=0.18\linewidth,height=3.55cm]{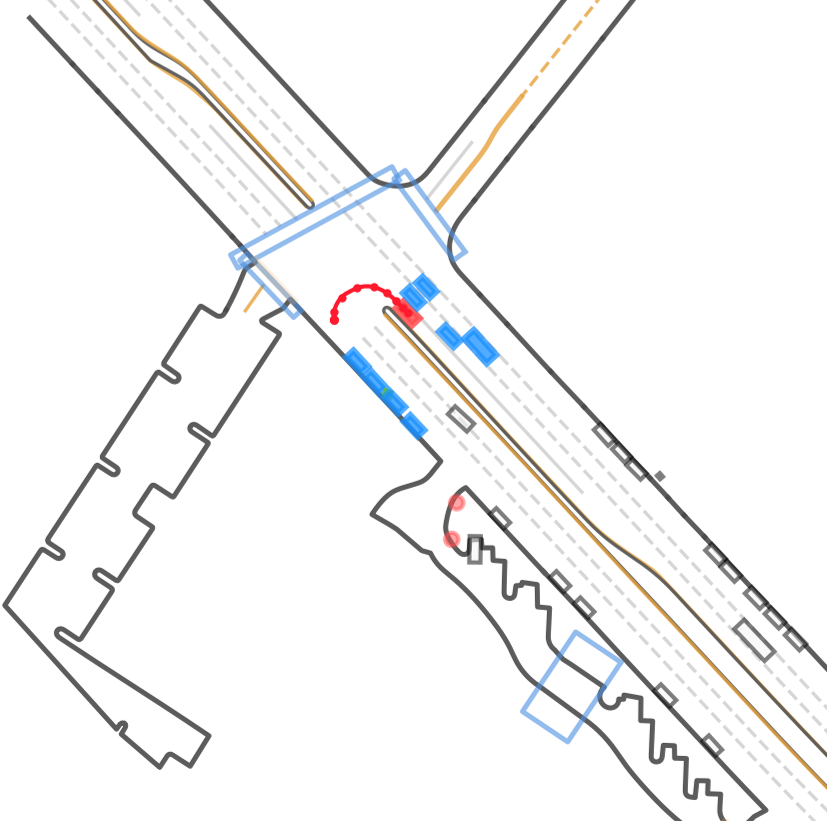}}} \hspace{0.5mm}
\subfloat{\fbox{\includegraphics[width=0.18\linewidth,height=3.55cm]{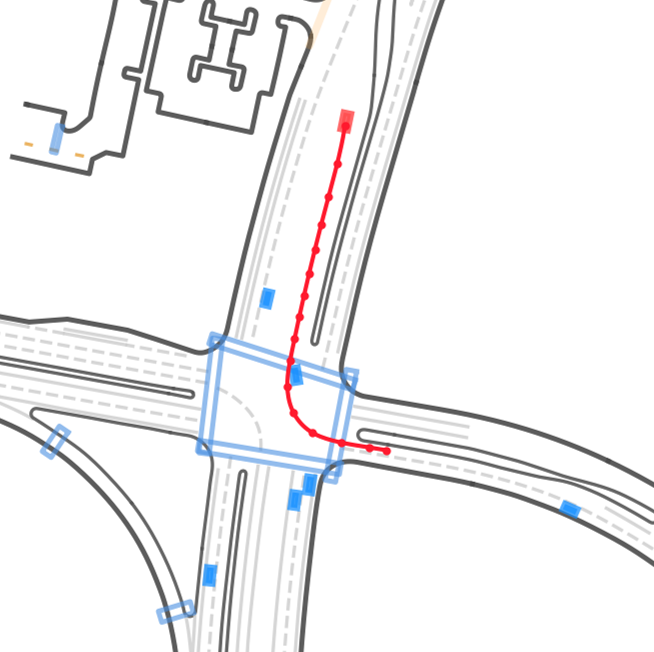}}} \hspace{0.5mm}
\subfloat{\fbox{\includegraphics[width=0.18\linewidth,height=3.55cm]{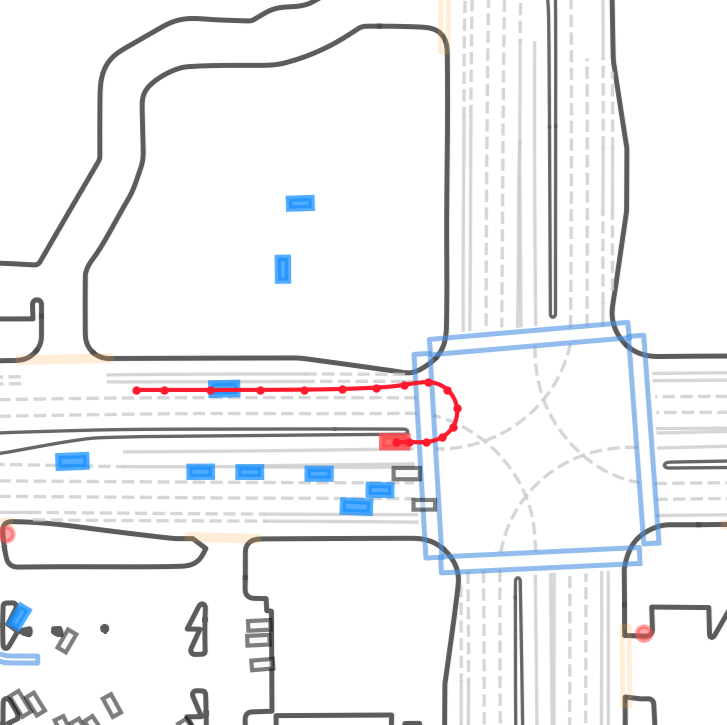}}} \hspace{0.5mm}
\subfloat{\fbox{\includegraphics[width=0.18\linewidth,height=3.55cm]{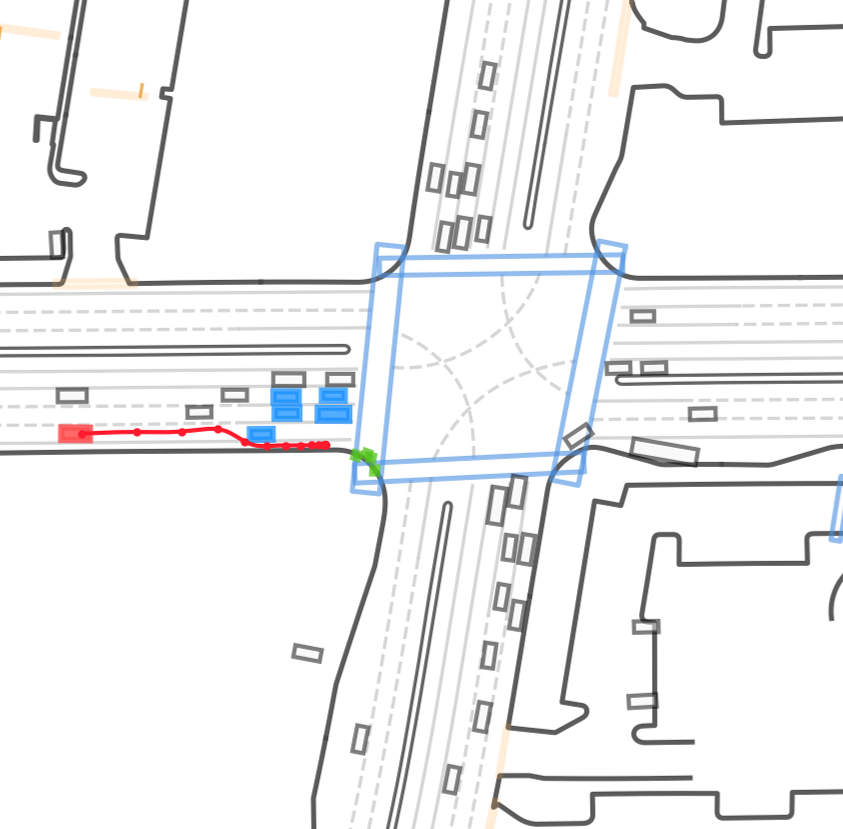}}} \\[-1mm]

 \caption{Qualitative evaluation of the proposed framework across diverse closed-loop scenarios. Colored solid lines represent the planned trajectory for the EV. For visual clarity, only the final position is displayed for each SA.
   }
    \label{Fig6}
\end{figure*}

From the experimental results, it can be observed that our framework demonstrates excellent performance in both safety and comfort. Our framework achieves the highest success rate in the WOMD datasets, representing improvements of 43.28\% compared to the second-best values and the second-highest success rate in the Argoverse 2 Motion Forecasting Dataset. The vehicle metrics also reach optimal or near-optimal values. Due to the high success rates, our framework achieves the longest progress in the simulator, with improvements of 38.90\% over the second-best values in the WOMD datasets. Furthermore, the position error relative to ground truth is also minimized. Fig.~\ref{Fig6} illustrates the qualitative experiment results in several representative scenarios. It can be observed that our method enables the EV to plan safe, collision-free trajectories across diverse urban scenarios.

Overall, our framework demonstrates a compelling balance of low collision risk, high route progression, and adequate passenger comfort in closed-loop scenarios, collectively exceeding the performance of prior baselines. The improvement in closed-loop safety suggests that conditioning trajectory optimization on sampled future trajectories enhances robustness against future interaction uncertainty.

 \begin{table*}[t]
          \centering
          \caption{comparison of the proposed framework with baselines in closed-loop testing.}
          \renewcommand{\arraystretch}{1.1}
          \subfloat[WOMD Scenario]{
            \begin{tabular}{c|cc|ccc|ccc @{\extracolsep{\fill}}}
            \toprule
            \multirow{2}{*}{Method} & Success $\uparrow$  & Progress $\uparrow$ & Acc. $\downarrow$ & Jerk $\downarrow$ & Lat. Acc. $\downarrow$ & \multicolumn{3}{c}{Position error (m) $\downarrow$}   \\
                  & (\%)   & (m) & (m/s$^2$) & (m/s$^3$) & (m/s$^2$) & @3$\,$s     & @5$\,$s   & @10$\,$s 
                   \\
            \midrule
            Vanilla IL & 60.00  & 8.52 & 6.73 & 98.48 & 5.73 & 11.91 & 23.30 & 50.18  \\
            IL + Prediction & 53.00  & 10.60 & 7.08 & 104.46 & 4.77 & 12.37 & 23.87 & 49.80   \\
            DIPP \cite{huang2023differentiable} & \underline{67.00}    &  \underline{57.94} &  \underline{0.97} &  \underline{4.14} &  \underline{1.84}  &  \underline{3.19}  & 7.10 &  \underline{18.83}   \\
            GameFormer \cite{Huang_2023_ICCV} & 45.00 & 41.68 & 1.19 &  8.27
            & 1.99 & 2.96 &  \underline{6.99} & 23.69  \\
            Ours & \textbf{96.00}  & \textbf{80.48} & \textbf{0.72} & \textbf{2.50} & \textbf{0.21} 
            & \textbf{1.13} & \textbf{2.18} & \textbf{3.99}  \\
            \midrule
            Human & $\slash$ &  $\slash$ & 0.74 & 4.30 & 0.11 & $\slash$ & $\slash$ & $\slash$   \\
            \bottomrule
            \end{tabular}%
            \label{tab:41}%
          }
        \hspace{1mm}
        \subfloat[Argoverse 2 Motion Forecasting Dataset Scenario]{

            \begin{tabular}{c|cc|ccc|ccc @{\extracolsep{\fill}}}
            \toprule
            \multirow{2}{*}{Method} & Success $\uparrow$ &  Progress $\uparrow$  & Acc. $\downarrow$ & Jerk $\downarrow$ & Lat. Acc. $\downarrow$ & \multicolumn{3}{c}{Position error (m) $\downarrow$}  \\
                  & (\%)  & (m) &(m/s$^2$) & (m/s$^3$) & (m/s$^2$) & @3$\,$s   & @5$\,$s   & @8$\,$s 
                   \\
            \midrule
            Vanilla IL & 82.00  & 5.41 & 7.05 & 112.11 & 3.40 & 13.21 & 16.76 & 18.50\\
            IL + Prediction & 82.00 & 9.92 & 5.57 & 92.71 & 2.57 & 10.89 & 14.06 & 16.13     \\
            DIPP \cite{huang2023differentiable} & \textbf{93.00}  & \textbf{30.72} & \underline{1.50} & 4.00 & \underline{1.28} & \underline{6.95} & \underline{11.95} & \underline{15.46} \\
            GameFormer\cite{Huang_2023_ICCV} & \underline{92.00} &   25.57 & 1.69 & \underline{3.78} & 1.95 & 13.79 & 16.94 & 18.17   \\
            Ours & \underline{92.00} & \underline{26.43} & \textbf{1.43} & \textbf{3.24} & \textbf{0.96} & \textbf{4.76} & \textbf{8.77} & \textbf{12.83} \\
            \midrule
            Human & $\slash$  & $\slash$ & $1.30$ & $17.23$ & $0.01$ & $\slash$ & $\slash$ & $\slash$    \\
            \bottomrule
            \end{tabular}%
            \label{tab:42}%
          }
        \label{tab:4}%
        \end{table*}%

\subsection{Ablation Studies}
The ablation study aims to validate the impact of individual components, key parameter configurations, and graph structure selections on the overall framework. The ablation experiments are organized into the following three parts:
\subsubsection{Effects of Individual Modules}
Three categories of ablation are conducted to assess the impact of removing specific components, compared with the baseline, which represents the complete implementation of the proposed framework:
\begin{enumerate}
    \item No Initialization on Planner: The initial control input to the planner is set to zeros rather than using the control initialization generated by the neural network.
    \item No Integrated Prediction: 
    The predicted trajectories of SAs are replaced by a constant turn rate and velocity model.
    \item No Learnable Cost Function: The cost weights in equation (\ref{cost}) are all set as constants rather than learnable parameters.
   \end{enumerate}
   
We conduct open-loop and closed-loop testing for ablation studies on the WOMD scenario, with the results included in Table \ref{tab:5}. In open-loop testing, we select collision rate, off-route rate, planning, and prediction errors as our primary metrics. Specifically, planning error is measured as the average over the 1$\,$s, 3$\,$s, and 5$\,$s prediction horizons, while prediction accuracy is assessed via minimum ADE. For closed-loop evaluation, we employ success rate, progress, and position error, which is measured as the average over the 3$\,$s, 5$\,$s, and 10$\,$s prediction horizons, as the main performance indicators.

The ablation results demonstrate that removing the planner's initial control input significantly degrades performance across both open-loop and closed-loop tests, substantially reducing safety and accuracy. Compared to the full framework, the collision rate and off-route rate increase by 380\% and 53.41\% in open-loop testing, respectively, while the average planning error rises by 335\%. In closed-loop testing, the success rate drops by 31\%, progress decreases by 26\%, and the average position error increases by 230\%. These results indicate that a reasonable initial control input enables the planner to converge to better solutions, whereas zero initialization fails to achieve convergence within 50 planning iterations, thereby compromising performance. Removing the prediction module also affects safety and accuracy. The average planning error increases in open-loop testing, along with higher collision rates. In closed-loop testing, the success rate declines. However, since the EV still receives network-generated initial control inputs, the performance degradation relative to the full framework remains moderate after planning refinement. Removing the learnable cost function exhibits a similar trend to removing the prediction module. The underlying reason is that the EV retains a relatively accurate initial control sequence, which partially compensates for the missing component.
 \begin{table*}[htbp]
          \centering
          \caption{Ablation study on the importance of each component in open-loop and closed-loop testing on the WOMD scenario.}
          \renewcommand{\arraystretch}{1.1}
          \subfloat[Open-loop testing]{
            \begin{tabular}{c|cccc @{\extracolsep{\fill}}}
            \toprule
            \multirow{2}{*}{Method} & Collision $\downarrow$  & Off-route $\downarrow$ & Position error (m) $\downarrow$ & Prediction error (m) $\downarrow$ \\
                  & (\%) & (Thre.=5$\,$m)(\%)  & Aver. (1$\,$s, 3$\,$s, 5$\,$s)   & minADE
                   \\
            \midrule
            No initialization on planner & $4.61$ & $1.35$ & $3.16$ & $0.70$ \\
            No integrated prediction & $1.86$ & $0.48$ & $1.23$ & $\slash$  \\
            No learnable cost function & $1.38$ & $0.84$ & $0.78$ & $0.70$  \\
            Ours & $0.96$ & $0.88$ & $0.75$ & $0.64$  \\
            \bottomrule
            \end{tabular}%
            \label{tab:51}%
          }
        \hspace{1mm}
        \subfloat[Closed-loop testing]{

            \begin{tabular}{c|cccc @{\extracolsep{\fill}}}
            \toprule
            \multirow{2}{*}{Method} & Success $\uparrow$ &  Progress $\uparrow$  & Position error (m) $\downarrow$   \\
                  & (\%)  & (m)  & Aver. (3$\,$s, 5$\,$s, 10$\,$s)  
                   \\
            \midrule
            No initialization on planner & 66.00 & 59.52 & 8.24  \\
            No integrated prediction & 91.00 & 80.42 & 2.52 \\
            No learnable cost function & 92.00 & 78.80 & 2.89 \\
            Ours & 96.00 & 80.48 & 2.43 \\
            \bottomrule
            \end{tabular}%
            \label{tab:52}%
          }
          \hspace{1mm}
        \begin{minipage}{0.5\linewidth}
        \footnotesize
        \vspace{0.5em}
        Aver.: average
        \end{minipage}
        \label{tab:5}%
        \end{table*}%
\subsubsection{Effects of Sample Size}
Generally, empirical CVaR-based tail-risk constraints require a large number of samples. However, our predictor models the joint distribution of all agents. In autonomous driving scenarios, the possible trajectories of vehicles are limited in number and concentrated in specific regions. Trajectories with excessively close endpoints are regarded as redundant predictions \cite{shi2022motion}. The diffusion model already concentrates mass on high-likelihood modes, and the tail-risk penalty focuses optimization on the most hazardous scenarios rather than the full distribution. Therefore, a small sample size can also meet the requirements. 

To investigate the impact of sample size $M$ on the results, we generate trajectories with varying values of $M$. We conduct open-loop and closed-loop testing on the WOMD scenario, with the results included in Table \ref{tab:6}. We select the collision rate and off-route rate for open-loop testing, as well as the success rate and progress for closed-loop testing, as our primary evaluation metrics.

From the experimental results, we observe that collision rates increase slightly when $M$ is small, as limited sample sizes fail to cover all possible scenarios, leading to collisions. Conversely, collision rates also rise modestly when $M$ is large. Although the risk tolerance $\delta$ mechanism filters out some extreme cases, a subset of retained extreme scenarios still influences the planner, causing marginally higher collision rates. Based on this analysis, we select $M=10$ for our experiments.

 \begin{table}[htbp]
          \centering
          \caption{Impact of sample size on Performance on the WOMD scenario.}
          \renewcommand{\arraystretch}{1.1}
          \subfloat[Open-loop testing]{
            \begin{tabular}{c|cc @{\extracolsep{\fill}}}
            \toprule
            \multirow{2}{*}{Method} & Collision $\downarrow$  & Off-route $\downarrow$  \\ & (\%) & (Thre.=5$\,$m)(\%)  \\
            \midrule
            $M=5$ & 1.09 & 0.87  \\
            $M=10$ & 0.96 & 0.88   \\
            $M=20$ & 1.30 & 0.86   \\
            \bottomrule
            \end{tabular}%
            \label{tab:61}%
          }
        \hspace{1mm}
        \subfloat[Closed-loop testing]{
            \begin{tabular}{c|cc @{\extracolsep{\fill}}}
            \toprule
            \multirow{2}{*}{Method} & Success $\uparrow$ &  Progress $\uparrow$  \\
                  & (\%)  & (m)  \\
            \midrule
             $M=5$ & 94.00 & 80.41  \\
            $M=10$ & 96.00 & 80.48   \\
            $M=20$ & 95.00 & 80.88   \\
            \bottomrule
            \end{tabular}%
            \label{tab:62}%
          }
        \label{tab:6}%
        \end{table}%
\subsubsection{Effects of Graph Structure}
To demonstrate that directed graph structure is more effective and efficient than the vectorized map, we conduct a comparative study. We maintain consistency across all components except for the graph encoder in the prediction network. For the vectorized map, we employ an MLP to encode local scene context for each agent. For the directed graph, we utilize GAT for encoding and employ it as a global scene context. We record the average training time per epoch throughout the training process as the basis for efficiency evaluation. Simultaneously, we record the prediction accuracy during testing, with minADE and minFDE serving as metrics for effectiveness evaluation. The experimental results are presented in Table \ref{tab:2}.

These results indicate that the directed graph structure provides more effective scene-context representations for sampled future trajectory generation, while simultaneously improving computational efficiency for downstream planning. Moreover, compared to the vectorized map structure, the directed graph structure exhibits lower computational resource occupancy. Under identical hardware configurations, the directed graph supports larger batch sizes. On our equipment, for instance, the directed graph accommodates a batch size of 512, whereas the vectorized map only supports 256.
\begin{table}[htbp]
          \centering
          \caption{Impact of Graph Structure on Prediction Performance and Training Efficiency on the WOMD scenario}
          \resizebox{1.0\linewidth}{!}{
            \begin{tabular}{c|ccc}
            \toprule
           Graph Structure & AverTrain  (s)* $\downarrow $ & minADE (m) $\downarrow $ & minFDE (m) $\downarrow $ \\
            \midrule
            Vectorized map  & 580     & 0.48    & 0.33   \\
            Directed graph & 176 ($69.7\%\downarrow$)   & 0.44 ($8.3\%\downarrow$)    & 0.28 ($15.6\%\downarrow$)    \\
            \bottomrule
            \end{tabular}%
          }\label{tab:2}%
          \hspace{1mm}
\begin{minipage}{0.95\linewidth}
\footnotesize
\vspace{0.5em}
\textit{EVerTrain*} represents the average training time per epoch throughout the training process.
\end{minipage}
\end{table}%

\subsection{Discussion}

Although the proposed framework demonstrates strong performance, several limitations remain. First, inference efficiency is constrained by diffusion sampling and iterative differentiable optimization. For future work, we will investigate reduced diffusion steps and lighter optimization settings for real-time deployment. Second, trajectory diversity remains limited in some scenarios due to the current loss design. To address this, future research includes the exploration of diversity-promoting objectives to better represent uncertainty while preserving endpoint accuracy.

\section{Conclusion}\label{sec:cond}
In this paper, we propose a sample-conditioned differentiable planning framework that explicitly incorporates diffusion-generated future trajectory samples into trajectory optimization, trained jointly on real-world driving data. The diffusion model takes concatenated observations as input, which comprise historical states, motion features, and directed graph-based scene context. The diffusion-based predictor models the conditional distribution of EV and SAs' motion given current observations. Through sampling, it generates future trajectory samples for SAs and an initial control sequence for the EV. These are fed into a sample-conditioned differentiable planner that produces safe driving trajectories via empirical CVaR-based tail-risk-constrained optimization. We validate the framework through open-loop and closed-loop testing on different large-scale urban driving datasets. The key advantage of the proposed framework lies in explicitly propagating sampled future uncertainty into differentiable trajectory optimization. Experimental results demonstrate that conditioning planning on sampled futures improves safety and robustness under uncertain traffic interactions, and our method outperforms baselines in safety, travel efficiency, and ride comfort. Future work will address inference latency from diffusion sampling and differentiable optimization, as well as enhance prediction diversity through refined loss design.


\begin{thebibliography}{10}

\bibitem{10122777}
K.~Yang, X.~Tang, J.~Li, H.~Wang, G.~Zhong, J.~Chen, and D.~Cao, ``{Uncertainties in Onboard Algorithms for Autonomous Vehicles: Challenges, Mitigation, and Perspectives},'' {\em IEEE Transactions on Intelligent Transportation Systems}, vol.~24, no.~9, pp.~8963--8987, 2023.

\bibitem{11077782}
W.~Shao, J.~Xu, Z.~Cao, H.~Wang, and J.~Li, ``{From Prediction to Planning: Comprehensive Uncertainty Management in Autonomous Driving},'' {\em IEEE Transactions on Intelligent Transportation Systems}, vol.~26, no.~10, pp.~16466--16480, 2025.

\bibitem{10122127}
S.~Teng, X.~Hu, P.~Deng, B.~Li, Y.~Li, Y.~Ai, D.~Yang, L.~Li, Z.~Xuanyuan, F.~Zhu, and L.~Chen, ``{Motion Planning for Autonomous Driving: The State of the Art and Future Perspectives},'' {\em IEEE Transactions on Intelligent Vehicles}, vol.~8, no.~6, pp.~3692--3711, 2023.

\bibitem{noh2017decision}
S.~Noh and K.~An, ``{Decision-making Framework for Automated Driving in Highway Environments},'' {\em IEEE Transactions on Intelligent Transportation Systems}, vol.~19, no.~1, pp.~58--71, 2017.

\bibitem{sankar2020adaptive}
G.~S. Sankar and K.~Han, ``{Adaptive Robust Game-theoretic Decision Making Strategy for Autonomous Vehicles in Highway},'' {\em IEEE Transactions on Vehicular Technology}, vol.~69, no.~12, pp.~14484--14493, 2020.

\bibitem{11081470}
B.~Ma, W.~Liu, and J.~Ma, ``{Trajectory Tree-Based Pairwise Game for Interactive Decision-Making and Motion Planning in Autonomous Driving},'' {\em IEEE Transactions on Vehicular Technology}, vol.~74, no.~12, pp.~18660--18672, 2025.

\bibitem{11494278}
Z.~Huang, T.~Li, S.~Shen, and J.~Ma, ``{Integrated Decision Making and Trajectory Planning for Autonomous Driving Under Multimodal Uncertainties: A Bayesian Game Approach},'' {\em IEEE Transactions on Vehicular Technology}, pp.~1--16, 2026.

\bibitem{10952911}
Z.~Guo, H.~Chen, F.~Xu, Y.~Hu, J.~Lin, and L.~Guo, ``{Uncertainty-aware Safe Trajectory Planner Based on Model Predictive Control for Autonomous Driving},'' {\em IEEE Transactions on Intelligent Transportation Systems}, vol.~26, no.~8, pp.~12068--12079, 2025.

\bibitem{9145612}
A.~Wang, A.~Jasour, and B.~C. Williams, ``{Non-Gaussian Chance-Constrained Trajectory Planning for Autonomous Vehicles Under Agent Uncertainty},'' {\em IEEE Robotics and Automation Letters}, vol.~5, no.~4, pp.~6041--6048, 2020.

\bibitem{Schwarting_2018}
W.~Schwarting, J.~Alonso-Mora, and D.~Rus, ``{Planning and Decision-making for Autonomous Vehicles},'' {\em Annual Review of Control, Robotics, and Autonomous Systems}, vol.~1, pp.~187--210, 2018.

\bibitem{Salzmann2020}
T.~Salzmann, B.~Ivanovic, P.~Chakravarty, and M.~Pavone, ``{Trajectron++: Dynamically-feasible Trajectory Forecasting with Heterogeneous Data},'' in {\em European Conference on Computer Vision (ECCV)}, pp.~683--700, 2020.

\bibitem{Lee_2017_CVPR}
N.~Lee, W.~Choi, P.~Vernaza, C.~B. Choy, P.~H.~S. Torr, and M.~Chandraker, ``{Desire: Distant Future Prediction in Dynamic Scenes with Interacting Agents},'' in {\em Proceedings of the IEEE Conference on Computer Vision and Pattern Recognition (CVPR)}, July 2017.

\bibitem{Gupta_2018_CVPR}
A.~Gupta, J.~Johnson, L.~Fei-Fei, S.~Savarese, and A.~Alahi, ``{Social Gan: Socially Acceptable Trajectories with Generative Adversarial Networks},'' in {\em Proceedings of the IEEE Conference on Computer Vision and Pattern Recognition (CVPR)}, June 2018.

\bibitem{Chai2018CoRL}
Y.~Chai, B.~Sapp, M.~Bansal, and D.~Anguelov, ``{Multipath: Multiple Probabilistic Anchor Trajectory Hypotheses for Behavior Prediction},'' in {\em Proceedings of the Conference on Robot Learning (CoRL)}, 2019.

\bibitem{pmlr-v155-zhao21b}
H.~Zhao, J.~Gao, T.~Lan, C.~Sun, B.~Sapp, B.~Varadarajan, Y.~Shen, Y.~Shen, Y.~Chai, C.~Schmid, C.~Li, and D.~Anguelov, ``{TNT: Target-driven Trajectory Prediction},'' in {\em Proceedings of the Conference on Robot Learning (CoRL)}, 2021.

\bibitem{9756903}
Y.~Huang, J.~Du, Z.~Yang, Z.~Zhou, L.~Zhang, and H.~Chen, ``{A Survey on Trajectory-prediction Methods for Autonomous Driving},'' {\em IEEE Transactions on Intelligent Vehicles}, vol.~7, no.~3, pp.~652--674, 2022.

\bibitem{Rhinehart_2018}
N.~Rhinehart, K.~M. Kitani, and P.~Vernaza, ``{R2P2: A Reparameterized Pushforward Policy for Diverse, Precise Generative Path Forecasting},'' in {\em European Conference on Computer Vision (ECCV)}, vol.~11217, pp.~47--62, 2018.

\bibitem{8953435}
O.~Makansi, E.~Ilg, O.~\c{C}i\c{c}ek, and T.~Brox, ``{Overcoming Limitations of Mixture Density Networks: A Sampling and Fitting Framework for Multimodal Future Prediction},'' in {\em IEEE/CVF Conference on Computer Vision and Pattern Recognition (CVPR)}, pp.~7137--7146, 2019.

\bibitem{Gu_2022_CVPR}
T.~Gu, G.~Chen, J.~Li, C.~Lin, Y.~Rao, J.~Zhou, and J.~Lu, ``{Stochastic Trajectory Prediction via Motion Indeterminacy Diffusion},'' in {\em Proceedings of the IEEE/CVF Conference on Computer Vision and Pattern Recognition (CVPR)}, pp.~17113--17122, June 2022.

\bibitem{Song_2021_ICLR}
Y.~Song, J.~Sohl-Dickstein, D.~P. Kingma, A.~Kumar, S.~Ermon, and B.~Poole, ``{Score-based Generative Modeling Through Stochastic Differential Equations},'' in {\em Proceedings of the International Conference on Learning Representations (ICLR)}, 2021.

\bibitem{Jiang_2023_CVPR}
C.~M. Jiang, A.~Cornman, C.~Park, B.~Sapp, Y.~Zhou, and D.~Anguelov, ``{MotionDiffuser: Controllable Multi-agent Motion Prediction Using Diffusion},'' in {\em Proceedings of the IEEE/CVF Conference on Computer Vision and Pattern Recognition (CVPR)}, pp.~9644--9653, June 2023.

\bibitem{Bo_2026_ICLR}
B.~Jiang, S.~Chen, H.~Gao, B.~Liao, Q.~Zhang, W.~Liu, and X.~Wang, ``{VADv2: End-to-end Vectorized Autonomous Driving via Probabilistic Planning},'' in {\em Proceedings of the International Conference on Learning Representations (ICLR)}, 2026.

\bibitem{goodfellow2014explaining}
I.~J. Goodfellow, J.~Shlens, and C.~Szegedy, ``{Explaining and Harnessing Adversarial Examples},'' {\em arXiv preprint arXiv:1412.6572}, 2014.

\bibitem{rudin2019stop}
C.~Rudin, ``{Stop Explaining Black Box Machine Learning Models for High Stakes Decisions and Use Interpretable Models Instead},'' {\em Nature Machine Intelligence}, vol.~1, no.~5, pp.~206--215, 2019.

\bibitem{9197560}
H.~Cui, T.~Nguyen, F.-C. Chou, T.-H. Lin, J.~Schneider, D.~Bradley, and N.~Djuric, ``{Deep Kinematic Models for Kinematically Feasible Vehicle Trajectory Predictions},'' in {\em IEEE International Conference on Robotics and Automation (ICRA)}, pp.~10563--10569, 2020.

\bibitem{10740461}
Z.~Xu and Y.~She, ``{LETO: Learning Constrained Visuomotor Policy with Differentiable Trajectory Optimization},'' {\em IEEE Transactions on Automation Science and Engineering}, vol.~22, pp.~8567--8578, 2025.

\bibitem{liu2024integrating}
W.~Liu, Y.~Song, C.~Meng, Z.~Huang, H.~Liu, C.~Lv, and J.~Ma, ``{Integrating decision-making into differentiable optimization guided learning for end-to-end planning of autonomous vehicles},'' {\em arXiv preprint arXiv:2412.01234}, 2024.

\bibitem{huang2023differentiable}
Z.~Huang, H.~Liu, J.~Wu, and C.~Lv, ``{Differentiable Integrated Motion Prediction and Planning with Learnable Cost Function for Autonomous Driving},'' {\em IEEE Transactions on Neural Networks and Learning Systems}, vol.~35, no.~11, pp.~15222--15236, 2023.

\bibitem{ettinger2021large}
S.~Ettinger, S.~Cheng, B.~Caine, C.~Liu, H.~Zhao, S.~Pradhan, Y.~Chai, B.~Sapp, C.~R. Qi, Y.~Zhou, {\em et~al.}, ``{Large Scale Interactive Motion Forecasting for Autonomous Driving: The Waymo Open Motion Dataset},'' in {\em Proceedings of the IEEE/CVF International Conference on Computer Vision (ICCV)}, pp.~9710--9719, 2021.

\bibitem{wilsonargoverse}
B.~Wilson, W.~Qi, T.~Agarwal, J.~Lambert, J.~Singh, S.~Khandelwal, B.~Pan, R.~Kumar, A.~Hartnett, J.~K. Pontes, {\em et~al.}, ``{Argoverse 2: Next Generation Datasets for Self-driving Perception and Forecasting},'' in {\em Advances in Neural Information Processing Systems}, 2021.

\bibitem{diff1}
C.~Chi, Z.~Xu, S.~Feng, E.~Cousineau, Y.~Du, B.~Burchfiel, R.~Tedrake, and S.~Song, ``{Diffusion Policy: Visuomotor Policy Learning via Action Diffusion},'' {\em The International Journal of Robotics Research}, vol.~44, no.~10-11, pp.~1684--1704, 2025.

\bibitem{song2019generative}
Y.~Song and S.~Ermon, ``{Generative Modeling by Estimating Gradients of the Data Distribution},'' {\em Advances in Neural Information Processing Systems}, vol.~32, 2019.

\bibitem{brown2020language}
T.~Brown, B.~Mann, N.~Ryder, M.~Subbiah, J.~D. Kaplan, P.~Dhariwal, A.~Neelakantan, P.~Shyam, G.~Sastry, A.~Askell, {\em et~al.}, ``{Language Models Are Few-shot Learners},'' {\em Advances in Neural Information Processing Systems}, vol.~33, pp.~1877--1901, 2020.

\bibitem{salimans2022progressive}
T.~Salimans and J.~Ho, ``{Progressive Distillation for Fast Sampling of Diffusion Models},'' {\em arXiv preprint arXiv:2202.00512}, 2022.

\bibitem{rajamani2011lateral}
R.~Rajamani, ``{Lateral Vehicle Dynamics},'' in {\em Vehicle Dynamics and Control}, pp.~15--46, Springer, 2011.

\bibitem{amos2018differentiable}
B.~Amos, I.~Jimenez, J.~Sacks, B.~Boots, and J.~Z. Kolter, ``{Differentiable Mpc for End-to-end Planning and Control},'' {\em Advances in Neural Information Processing Systems}, vol.~31, 2018.

\bibitem{charnes1959chance}
A.~Charnes and W.~W. Cooper, ``{Chance-constrained Programming},'' {\em Management Science}, vol.~6, no.~1, pp.~73--79, 1959.

\bibitem{8767973}
A.~Hakobyan, G.~C. Kim, and I.~Yang, ``{Risk-Aware Motion Planning and Control Using CVaR-Constrained Optimization},'' {\em IEEE Robotics and Automation Letters}, vol.~4, no.~4, pp.~3924--3931, 2019.

\bibitem{rockafellar2000optimization}
R.~T. Rockafellar, S.~Uryasev, {\em et~al.}, ``Optimization of conditional value-at-risk,'' {\em Journal of Risk}, vol.~2, pp.~21--42, 2000.

\bibitem{Huang_2023_ICCV}
Z.~Huang, H.~Liu, and C.~Lv, ``{GameFormer: Game-theoretic Modeling and Learning of Transformer-based Interactive Prediction and Planning for Autonomous Driving},'' in {\em Proceedings of the IEEE/CVF International Conference on Computer Vision (ICCV)}, pp.~3903--3913, October 2023.

\bibitem{songdenoising}
J.~Song, C.~Meng, and S.~Ermon, ``{Denoising Diffusion Implicit Models},'' in {\em Proceedings of the International Conference on Learning Representations (ICLR)}, 2021.

\bibitem{shi2022motion}
S.~Shi, L.~Jiang, D.~Dai, and B.~Schiele, ``{Motion Transformer with Global Intention Localization and Local Movement Refinement},'' {\em Advances in Neural Information Processing Systems}, 2022.

\end{thebibliography}
\end{document}